\PassOptionsToPackage{capitalize}{cleveref}
\documentclass[]{fairmeta}

\usepackage{amsmath}
\usepackage{amssymb}
\usepackage{xspace}
\usepackage[shortlabels,inline]{enumitem}
\usepackage{wrapfig} %
\usepackage{fontawesome5} %
\newcommand{\hficon}{\raisebox{-0.18em}{\includegraphics[height=1.05em]{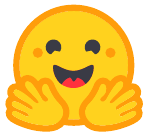}}}
\DeclareFontShape{T1}{optimistic}{b}{n}{<-> s * [0.88] assets/optimistic}{}
\DeclareFontShape{T1}{optimistic}{b}{sc}{<-> s * [0.88] assets/optimistic}{}
{\fontencoding{T1}\fontfamily{cmvtt}\selectfont}%
\DeclareFontShape{T1}{cmvtt}{b}{n}{<->ssub * cmvtt/m/n}{}
\setlist[itemize]{noitemsep,leftmargin=*,topsep=0em}
\setlist[enumerate]{noitemsep,leftmargin=*,topsep=0em}
\makeatletter
\DeclareRobustCommand\onedot{\futurelet\@let@token\@onedot}
\def\@onedot{\ifx\@let@token.\else.\null\fi\xspace}
 
\def\ie{\emph{i.e}\onedot}

\makeatother

\usepackage{booktabs}
\newcommand{\bench}{\textsc{TuringSpace}\xspace} %
\newcommand{\benchname}{\textsc{TuringSpace-Bench}\xspace} %
\newcommand{\gymname}{\textsc{TuringSpace-Gym}\xspace} %
\newcommand{\modelname}{\textsc{TuringSpace-9B}\xspace}
\usepackage{xurl} %
\usepackage{pifont}
\definecolor{cmarkgreen}{cmyk}{0.91,0,0.88,0.12} %
\definecolor{xmarkred}{cmyk}{0,0.89,0.94,0.28}    %
\newcommand{\cmark}{\textcolor{cmarkgreen}{\ding{51}}}   %
\newcommand{\xmark}{\textcolor{xmarkred}{\ding{55}}}   %
\usepackage{calc}

\usepackage{colortbl}
\definecolor{keyblue}{HTML}{5A85E1}
\definecolor{keybg}{HTML}{F1F7FF}
\definecolor{hlrow}{HTML}{B4FFB4}
\usepackage[most]{tcolorbox}
\newtcolorbox{rqbox}{enhanced, colback=keybg, colframe=keyblue, boxrule=0.6pt, arc=1.5pt, left=6pt, right=6pt, top=3pt, bottom=3pt, borderline west={3pt}{0pt}{keyblue}, before skip=5pt, after skip=5pt}

\newcommand{\passk}{pass\textasciicircum{}3\xspace} %
\newcommand{\mlogo}[1]{\makebox[1.15em][l]{\raisebox{-0.17em}{\includegraphics[height=0.95em,width=0.95em,keepaspectratio]{figures/logos/#1.pdf}}}} %

\title{TuringSpace: Grounding Spatial Intelligence in Action with Interactive CAPTCHAs}

\author[1,*]{Xuwei~Ding}
\author[2,*]{Skylar~Zhai}
\author[3]{Linxin~Song}
\author[4]{Jiarui~Feng}
\author[4]{Dongqi~Fu}
\author[5]{Huajun~Xi}
\author[6]{Junnan~Li}
\author[2]{Jingcheng~Liang}
\author[3]{Jieyu~Zhao}
\author[4]{Yinglong~Xia}
\author[4]{Hong~Li}
\author[5]{Jian~Kang}

\affiliation[1]{University of Pennsylvania}
\affiliation[2]{University of Minnesota}
\affiliation[3]{University of Southern California}
\makeatletter\addtolist[4]{Meta}{\affiliationlist}{\affiliationformat}{,\protect\newline}\makeatother
\affiliation[5]{MBZUAI}
\affiliation[6]{University of Wisconsin--Madison}

\abstract{
Humans act in the visual world by inferring the 3D layout behind a 2D view, predicting how an action changes it, and correcting themselves after seeing the result.
Can multimodal large language models (MLLMs) do the same? 
CAPTCHAs offer a natural test, since humans solve them through this very loop, clicking and dragging on the screen and adjusting after each result.
Yet existing CAPTCHA benchmarks are largely solved and rarely spatial, while spatial benchmarks rarely require action.
We therefore present TuringSpace, a synthetic web environment of 125 interactive CAPTCHAs that span five spatial abilities and must be solved through clicks and drags on the screen.
Every task is a program that generates new instances with a checker and an oracle, so the tasks also form a scalable training gym.
Humans solve 94.1\% of the tasks, whereas the best model, GPT-6 Astra, still fails 25.9\% of them, and the best open-weight model, Kimi K3, fails 68.0\%.
Our further analysis shows that most failures come from misreading the screenshot, and that model reasoning rarely corrects a wrong belief but hallucinates evidence to defend it.
To test whether such spatial abilities can be learned, we fine-tune \texttt{Qwen3.5-9B} on this gym.
Its success rate rises from 1.6\% to 33.1\%, above Kimi K3, and it improves on all seven out-of-distribution benchmarks, including a 9.9-point gain on the real-world indoor videos of VSI-Bench.
These results show that grounding spatial intelligence in screen actions exposes what MLLMs lack and offers a scalable way to teach it, with gains that transfer to real-world scenes.
We will release TuringSpace to support further study.
}

\date{\today}

\newcommand{\projectpageurl}{https://xuweid.com/turingspace/}
\newcommand{\hfurl}{https://huggingface.co/datasets/zhaihaotian/TuringSpace-Gym} %
\newcommand{\githuburl}{https://github.com/os-cua-framework/TuringSpace}
\makeatletter
\g@addto@macro\metadatalist{\par{\small\sffamily\bfseries
  \href{\projectpageurl}{\textcolor{metafg}{\faGlobe}~Project Page}\hspace{1.5em}%
  \href{\githuburl}{\textcolor{metafg}{\faGithub}~GitHub}\hspace{1.5em}%
  \href{\hfurl}{\hficon~Hugging Face}
  }}
\makeatother

\begin{document}

\maketitle
\begingroup
\renewcommand{\thefootnote}{\fnsymbol{footnote}}
\footnotetext[1]{Equal contribution}
\endgroup

\section{Introduction}
\label{sec:intro}

\suppressfloats[t]
\begin{wrapfigure}{r}{0.45\linewidth}
  \vspace{-1.2em}
  \centering
  \includegraphics[width=\linewidth]{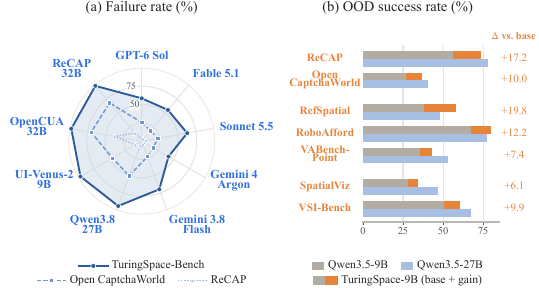}
  \caption{(a) Failure rate (\%) of nine MLLMs on \benchname{} and two CAPTCHA benchmarks; every model fails more often on \benchname{} than on both.
  (b) Success rate (\%) of Qwen3.5-9B, Qwen3.5-27B, and \modelname{} on seven out-of-distribution benchmarks; orange marks the gain over the base model.}
  \label{fig:radar}
  \vspace{-1.6em}
\end{wrapfigure}

Spatial understanding underlies how humans act in the visual world.
People can easily recover 3D structure from a flat view, mentally simulate the effect of an intervention, and refine each movement through the visual feedback it produces~\citep{li2026stare,li2026spacecast}.
Multimodal large language models (MLLMs)~\citep{anthropic2026opus55card,openai2026gpt6,google2026gemini4argon,qwen2026qwen38} now perform well on 2D perception but remain far behind humans in 3D spatial understanding, such as imagining a scene from another viewpoint or picturing an object after it rotates~\citep{cai2025holisticspatial,yang2026mmsibench,ramakrishnan2025space,wang2026mindcube,wang2026spatialviz}.
CAPTCHAs (Completely Automated Public Turing tests to tell Computers and Humans Apart) need exactly such gaps between humans and machines~\citep{vonahn2003captcha,vonahn2008recaptcha}, so they must move to new abilities once automated agents master the old ones.
\begin{figure*}[t]
  \centering
  \includegraphics[width=\linewidth]{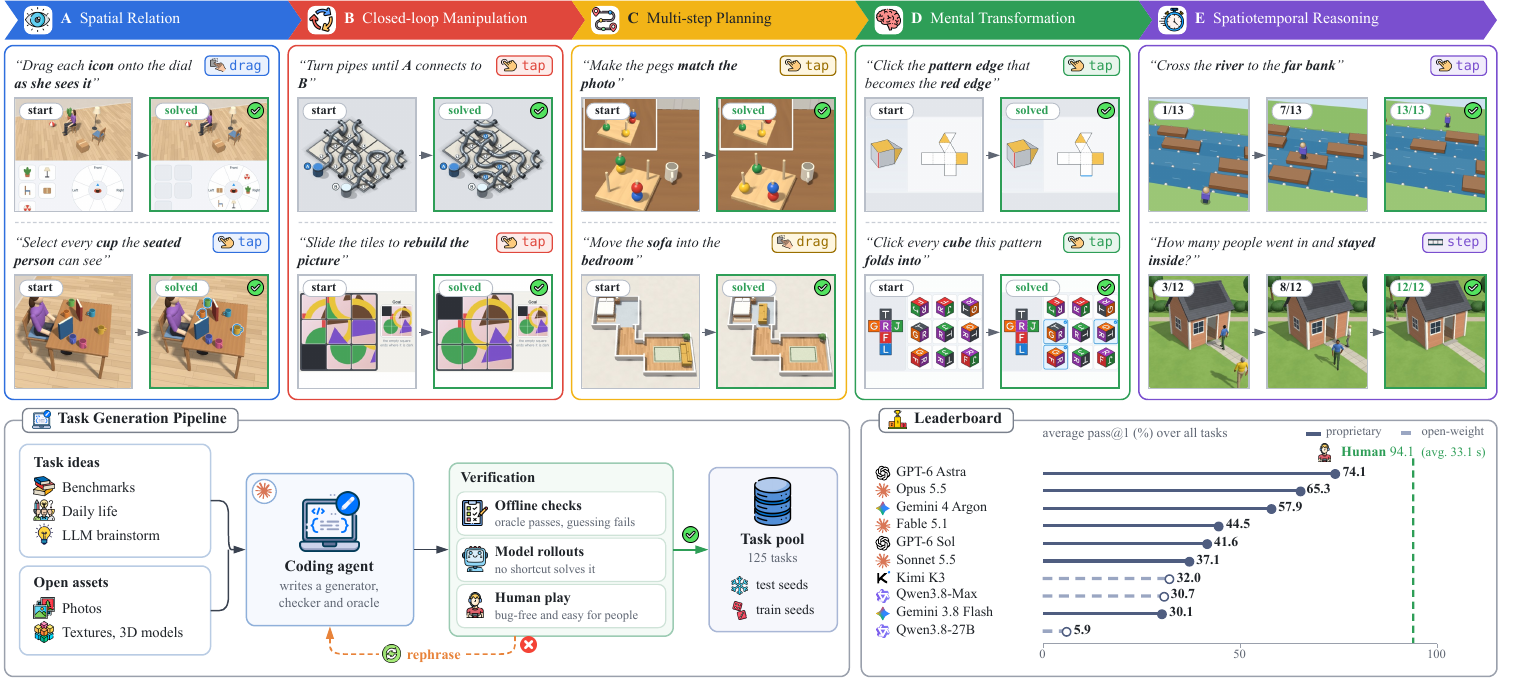}
  \caption{Overview of \bench{}.
  \textit{Top}: two of the 25 tasks in each of the five categories, with the instruction, start state, solved state, and a tag for how the player acts (tap, drag, or step); spatiotemporal tasks show three frames of their clip.
  \textit{Bottom left}: how a task is built and reviewed.
  \textit{Bottom right}: average pass@1 of the evaluated models over three seeds, against the human score.}
  \label{fig:teaser}
\end{figure*}

However, computer-use agents (CUAs) built on MLLM backbones~\citep{xie2024osworld,zhou2024webarena,anthropic2026opus55card,openai2026gpt6,qin2025uitars} already solve conventional CAPTCHAs with high success rates~\citep{kharlamova2026spatialcaptcha,teoh2025halligan,chen2026recap} (\cref{fig:radar}a).
These puzzles require only simple GUI actions on a single screen, such as typing distorted text, clicking matching image tiles, or dragging a slider.
A more promising and challenging direction shifts to building CAPTCHAs on spatial understanding, and recent work has already turned to it.
For example, Spatial CAPTCHA~\citep{kharlamova2026spatialcaptcha} asks which cube a flat net folds into or what a person in the scene can see, and shows a wide gap between humans and MLLMs.
Spatial understanding thus gives CAPTCHAs the gap between humans and machines that conventional puzzles have lost.

Despite this progress, current spatial CAPTCHAs~\citep{kharlamova2026spatialcaptcha,liu2026nextgencaptchas}, like most spatial benchmarks~\citep{wang2026mindcube,wang2026spatialviz,yang2025thinkinginspace,jia2026omnispatial}, are largely evaluated with multiple-choice questions.
Models could guess or rule out options in such a format, so a correct answer may not show that they understand the scene.
Where actions are allowed, they become symbolic choices: spatial environments accept only text commands such as ``turn left''~\citep{mayer2025ivispar,wang2026visgym,gao2026spatialworld}, and multiple CAPTCHA benchmarks expose the page's document object model (DOM), so the agent merely picks a labeled element~\citep{luo2025opencaptchaworld,wu2026mcabench}.
Some CAPTCHA benchmarks do support real GUI interaction~\citep{chen2026recap}, but their puzzles have largely been solved by CUAs~\citep{venus2026uivenus2}.
As a result, existing CAPTCHA benchmarks cannot serve as a good testbed for spatial reasoning through a sequence of actions.

This limitation matters because spatial reasoning in human cognition is not a one-shot judgment but a loop.
People read the 3D scene, predict the outcome of a move, act, and adjust after seeing the result.
Actions express every step of this loop far more richly than options~\citep{hong2026esibenchembodiedspatialintelligence}: a click specifies a location, a drag specifies a direction and a distance, and a sequence of actions specifies how the scene should change.
Consider an interactive spatial CAPTCHA that asks the player to move a sofa around a corridor corner.
When the next screenshot shows the sofa blocked, the player moves it back and rotates it again.
Solving it thus takes several rounds of observing and adjusting.
This loop thus grounds spatial understanding in action: each step reflects the agent's current belief about the scene, and each new observation can correct that belief.

To test this loop, we introduce \bench{}, a synthetic web environment of interactive spatial puzzles.
We evaluate models on \benchname{}, which contains 125 interactive CAPTCHAs that cover five spatial abilities (\cref{fig:teaser}, top).
Each task is solved through computer use, where an agent reads screenshots of a real browser and acts with the mouse and keyboard.
To guarantee their quality, all tasks are validated by human players.
Since a deployed CAPTCHA must synthesize a new puzzle for every visitor so that its answers cannot be memorized, each of our tasks is a program rather than a fixed puzzle.
The program can synthesize unlimited fresh instances from random seeds and verify the final state of each.
Replaying the solution that each task program computes yields a trajectory that pairs every action with its visual outcome, so the same tasks also form a training environment, \gymname{}.
To test whether such trajectories can teach models the spatial understanding they lack, we fine-tune \texttt{Qwen3.5-9B}~\citep{qwen2026qwen35} on them.

Our evaluation on \benchname{} reveals a large gap between humans and MLLMs.
The strongest model, GPT-6 Astra, solves 74.1\% of the tasks, compared with 94.1\% for humans, and no specialized CUA solves more than 3.5\%.
Their failures are systematic: for the strongest models, most come from misreading the screen, even when a screenshot shows the feedback of their own action.
Stronger models also do not think more than weaker ones, but they spend more of their thinking on hard tasks.
More interestingly, the reasoning of open-weight models rarely corrects a wrong belief formed early and instead hallucinates evidence to defend it.
To test whether spatial understanding could be learned from interaction, fine-tuning \texttt{Qwen3.5-9B} on 55,000 verified trajectories generated by \gymname{} raises its pass rate from 1.6\% to 33.1\%, above Kimi K3 and Qwen3.8-Max.
The resulting \modelname{} also improves on other CAPTCHA, spatial understanding, and robotic pointing benchmarks (\cref{fig:radar}b), even gaining 9.9 points on real-world videos from VSI-Bench~\citep{yang2025thinkinginspace}.
These results suggest that spatial understanding grounded in screen actions transfers to real-world scenes.
Therefore, interactive GUI environments may offer a scalable way to train the spatial abilities of MLLMs.
Our \mbox{contributions} are as follows.
\begin{itemize}
  \item \textbf{\benchname{}.} It contains 125 interactive spatial CAPTCHAs that cover five spatial abilities. Frontier MLLMs trail humans on these tasks.
  \item \textbf{\gymname{}.} Every task program synthesizes unlimited fresh instances, and replaying their oracle solutions yields verified ground-truth trajectories for training.
  \item \textbf{Interactive spatial training.} \modelname{} trained in \gymname{} improves on \benchname{} and transfers to CAPTCHA solving, spatial understanding, and robotic pointing and affordance.
\end{itemize}

\section{Related Work}
\label{sec:related}

\noindent\textbf{CAPTCHA benchmarks and solvers.}
CAPTCHAs tell humans and machines apart with puzzles such as distorted text, image grids, sliders, and icon matching~\citep{vonahn2003captcha,vonahn2008recaptcha}.
Recent CAPTCHA benchmarks are built from these classic puzzles~\citep{luo2025opencaptchaworld,teoh2025halligan,wu2026mcabench}, which programs generate in bulk.
The same generators supply training data, on which ReCAP~\citep{chen2026recap}, CaptchaArena~\citep{zhang2026captchaarena}, and CaptchaMind~\citep{wang2026captchamind} train agents that solve most of these puzzles~\citep{teoh2025halligan,chen2026recap}.
Some benchmarks also expose the page's DOM, so the agent only selects among labeled elements~\citep{luo2025opencaptchaworld,wu2026mcabench}.
Of the works closest to ours, Spatial CAPTCHA~\citep{kharlamova2026spatialcaptcha} tests perspective taking and mental rotation but only through single-choice questions.
Next-Gen CAPTCHAs~\citep{liu2026nextgencaptchas} builds dynamic puzzles that MLLMs rarely pass, but several rely on animation or timed clicks that screenshots cannot convey.
In contrast, \bench{} follows the standard CUA protocol~\citep{xie2024osworld} and has the agent solve spatial puzzles by clicking and dragging on screenshots without relying on timing or motion, so a failure exposes a genuine capability gap.

\noindent\textbf{Spatial understanding in MLLMs.}
Multiple benchmarks evaluate frontier MLLMs on spatial abilities such as 3D relation understanding, perspective taking, mental rotation, and visual simulation~\citep{yang2025thinkinginspace,ramakrishnan2025space,ma2025threedsrbench,wang2025spatial457,li2025viewspatial,wang2026mindcube,jia2026omnispatial,wang2026spatialviz,li2026stare,yang2026mmsibench}, revealing that these models still fall far behind humans.
However, almost all of these benchmarks test the abilities through multiple-choice questions about static images or videos, rather than by letting the model act.
A few recent works test spatial abilities in an interactive way, either in puzzle games~\citep{mayer2025ivispar,wang2026visgym} or in simulated embodied scenes~\citep{gao2026spatialworld,hong2026esibenchembodiedspatialintelligence}.
Yet the model in these works acts only through some simple text commands.
\bench{} instead poses spatial problems as interactive CAPTCHAs, where the model clicks and drags on the screen like a human and adjusts to the visual feedback of each action.

\noindent\textbf{Computer-use agents.}
Computer-use agents (CUAs) use an MLLM as their backbone to operate software from screenshots with mouse and keyboard actions~\citep{anthropic2026opus55card,openai2026gpt6,google2026gemini4argon,qwen2026qwen38,minimax2026m3,qin2025uitars,wang2025uitars2,wang2025opencua,xue2026evocua,zhou2026qwenuiagent,venus2026uivenus2}.
Their benchmarks focus on completing everyday tasks on desktops, websites, and phones successfully and safely~\citep{xie2024osworld,yuan2026osworld2,zhou2024webarena,xue2025onlinemind2web,rawles2025androidworld,bonatti2024windowsagentarena,li2025screenspotpro,ding2026blindspotagentsafety}.
These benchmarks exclude CAPTCHAs~\citep{luo2025opencaptchaworld}, although a CUA automating tasks on real websites must pass them.
Recent CUAs such as UI-Venus-2~\citep{venus2026uivenus2} have started to train on CAPTCHAs.
CUAs can also take on spatial tasks beyond software, since \citet{aggarwal2026cuaembodied} wrap embodied tasks such as navigation and 3D segmentation into applications where a general CUA outperforms specialized embodied models.
\bench{} provides verifiable spatial CAPTCHAs for evaluating and training MLLMs as CUAs, and the trained MLLM improves on other CAPTCHA, spatial understanding, and robotic pointing benchmarks.

\section{\bench{}: Benchmarking and Improving Spatial Understanding in MLLMs}
\label{sec:bench}

\bench{} is a suite that benchmarks and improves the spatial understanding of MLLMs through actions, built on 125 seed tasks that generate interactive spatial CAPTCHAs (\cref{fig:teaser}).
It is formed by \benchname{}, which evaluates MLLMs on three held-out instances per task, and \gymname{}, which generates unlimited training instances with disjoint assets.

\subsection{Design principles and task categories}
\label{sec:bench:principles}

\noindent\textbf{Design principles.}
Three principles guide the design of tasks in \bench{}.
First, each task is easy for humans and requires spatial understanding: it needs no expert knowledge, and its answer depends on the layout of the scene.
Second, it is interactive and fully observable: the player acts on each screenshot and sees the result in the next, with no time limit, and every cue is on screen.
Third, it is a program, not a fixed puzzle, so every seed yields a new instance with an automatic checker and an oracle solution.

\noindent\textbf{Task categories.}
The five categories differ in the spatial ability they test and in how the player interacts with the page (\cref{fig:teaser}, top).
The first two ask the player to mark or select an answer in a fixed scene.
\emph{Spatial relation understanding} recovers 3D relations such as occlusion, containment, support, and what another person can see, while \emph{mental transformation} depends on an operation that the page never shows, such as rotating, folding, cutting, or viewing an object from another position in the mind.
The next two require a sequence of actions that change the scene.
In \emph{closed-loop visual manipulation}, the player rotates, moves, or aligns objects and corrects each action after seeing its result.
In \emph{multi-step planning and simulation}, by contrast, a single result does not reveal whether the player is on track, so the player must plan ahead to avoid dead ends or predict how a process such as rolling, falling, or a reflected light beam unfolds.
Finally, \emph{spatiotemporal reasoning} adds time: the player advances up to 16 frames one at a time and must track or predict moving objects and people, or act in a world that keeps moving.
Together, the categories progress from reading 3D structure in one view, to transforming it in the mind, to acting on it with and without immediate feedback, and to following it over time.

\subsection{Building seed tasks}
\label{sec:bench:build}

\noindent\textbf{Task ideas and implementation.}
Task ideas are adapted from spatial benchmarks and cognitive tests, drawn from everyday spatial activities, or brainstormed with an LLM (\cref{fig:teaser}, bottom left).
For each idea, the coding agent Claude Code~\citep{anthropic2026claudecode} with Claude Opus 5.5~\citep{anthropic2026opus55card} writes a task program on a shared page runtime.
Each program turns a seed into a complete puzzle: an interactive page with an instruction, a checker for the final state, and an oracle solution.
A manually collected asset pool $\mathcal{M}$ holds openly licensed photos, textures, and 3D models from Open Images~\citep{kuznetsova2020openimages}, COCO~\citep{lin2014coco}, Poly Haven~\citep{polyhaven}, and ambientCG~\citep{ambientcg}.
Task programs draw on $\mathcal{M}$ to vary the objects and appearance of each instance.

\noindent\textbf{Review loop.}
Every task in \bench{} goes through a multi-round review to guarantee its quality (\cref{fig:teaser}, bottom left).
Each round goes through three checks, and any failed check returns to the coding agent as feedback for revision.
First, each task is rolled out on 100 seeds for programmatic checks.
Every instance must render without bugs, regenerate identically from its seed, and be solved by its oracle solution, while doing nothing or random clicking must almost never pass.
Second, Claude Sonnet 5.5~\citep{anthropic2026sonnet55} plays the task once, and the authors review its trajectory for shortcuts, \ie ways to pass without spatial understanding.
Third, two of the authors play the task and confirm that it has no bugs and that a human can solve it easily without ambiguity.
Together, these checks ensure that every task is correct, resists guessing and shortcuts, and remains easy for humans.

\subsection{From seed tasks to unlimited instances}
\label{sec:bench:instances}

\noindent\textbf{Seed tasks and instances.}
A seed task $\tau$ is a generator $G_\tau$ that maps a seed $z$ and an asset pool $\mathcal{M}$ to an instance that bundles the page, instruction, checker, and oracle solution of one puzzle.
\begin{equation}
  x = G_\tau(z, \mathcal{M}) = (s_0,\ q,\ C,\ a^\star_{1:L}).
  \label{eq:instance}
\end{equation}
Here, $s_0$ is the initial page state, $q$ is the instruction, and $C$ is the checker, which returns 1 if the final state solves the task and 0 otherwise.
The oracle solution $a^\star_{1:L}$ is a sequence of $L$ actions from $s_0$ to a final state $s_L$ with ${C(s_L) = 1}$. A new seed changes the layout, objects, viewpoint, and its solution, while the spatial ability that $\tau$ tests stays the same.

\noindent\textbf{Benchmark and training splits.}
The benchmark and the training data apply the same generators to different seeds and disjoint asset pools,
\begin{equation}
  \begin{aligned}
    \mathcal{D}_\text{test}  &= \{\, G_\tau(z, \mathcal{M}_\text{test})  \mid \tau \in \mathcal{T},\ z \in \{36, 43, 50\} \,\}, \\
    \mathcal{D}_\text{train} &= \{\, G_\tau(z, \mathcal{M}_\text{train}) \mid \tau \in \mathcal{T},\ z \in \{1, \dots, N\} \,\},
  \end{aligned}
  \label{eq:splits}
\end{equation}
where $\mathcal{T}$ is the set of tasks, and the three test seeds were chosen at random.
To prevent any data leakage, we separate the training data from the benchmark in three ways.
First, the training data skip the three test seeds, so no training instance shares a seed with a benchmark instance.
Second, the asset pools are disjoint, $\mathcal{M}_\text{train} \cap \mathcal{M}_\text{test} = \varnothing$, so no benchmark asset appears in training.
Third, after the rollouts, we reject any training instance whose spatial configuration has the same hash as that of a benchmark instance.

\noindent\textbf{Oracle trajectories.}
Replaying the oracle solution on the page yields a trajectory $\xi = (q, o_0, a^\star_1, o_1, \dots, a^\star_L, o_L)$, where $o_l$ is the screenshot of the state $s_l$ after the action $a^\star_l$.
We replay it with the clicks and drags of the evaluation harness and keep $\xi$ only if $C(s_L) = 1$, so each seed task turns into an unlimited stream of verified demonstrations in the action format of evaluation.

\begin{table*}[!t]
  \newcommand{\hlM}[1]{\multicolumn{1}{>{\columncolor{hlrow}[\tabcolsep][\tabcolsep]}c@{\hspace{7pt}}}{#1}}
  \newcommand{\std}[1]{\,{\scriptsize$\pm$#1}}
  \newcommand{\hlR}[1]{\multicolumn{1}{>{\columncolor{hlrow}[\tabcolsep][0pt]}w{c}{30.5pt}@{}}{#1}}
  \centering
  \small
  \setlength{\tabcolsep}{1.8pt}
  \caption{\bench{} results of humans, 16 MLLMs in three groups, and our fine-tuned model, in \%.
  Pass@1 is averaged over the three test seeds with its standard deviation, and \passk{} is the share of tasks solved on all three.
  Humans take 33.1\,s per task on average.
  Green cells mark \modelname{} and the three models that it surpasses in both metrics, and the best score among the evaluated models is in bold.}
  \label{tab:main}
  \resizebox{\linewidth}{!}{%
  \begin{tabular}{@{}lw{c}{30.5pt}w{c}{30.5pt}w{c}{30.5pt}w{c}{30.5pt}w{c}{30.5pt}c@{\hspace{7pt}}w{c}{30.5pt}w{c}{30.5pt}w{c}{30.5pt}w{c}{30.5pt}w{c}{30.5pt}w{c}{30.5pt}@{}}
    \toprule
    & \multicolumn{6}{c}{Pass@1} & \multicolumn{6}{c}{Pass\textasciicircum{}3} \\
    \cmidrule(lr){2-7} \cmidrule(l){8-13}
    \begin{tabular}[t]{@{}l@{}}\\Model\end{tabular} & {\fontsize{7}{8.5}\selectfont\begin{tabular}[t]{@{}c@{}}Spatial\\relation\end{tabular}} & {\fontsize{7}{8.5}\selectfont\begin{tabular}[t]{@{}c@{}}Mental\\transf.\end{tabular}} & {\fontsize{7}{8.5}\selectfont\begin{tabular}[t]{@{}c@{}}Closed-\\loop\end{tabular}} & {\fontsize{7}{8.5}\selectfont\begin{tabular}[t]{@{}c@{}}Multi-step\\planning\end{tabular}} & {\fontsize{7}{8.5}\selectfont\begin{tabular}[t]{@{}c@{}}Spatio-\\temporal\end{tabular}} & {\fontsize{7}{8.5}\selectfont\begin{tabular}[t]{@{}c@{}}All\\tasks\end{tabular}} & {\fontsize{7}{8.5}\selectfont\begin{tabular}[t]{@{}c@{}}Spatial\\relation\end{tabular}} & {\fontsize{7}{8.5}\selectfont\begin{tabular}[t]{@{}c@{}}Mental\\transf.\end{tabular}} & {\fontsize{7}{8.5}\selectfont\begin{tabular}[t]{@{}c@{}}Closed-\\loop\end{tabular}} & {\fontsize{7}{8.5}\selectfont\begin{tabular}[t]{@{}c@{}}Multi-step\\planning\end{tabular}} & {\fontsize{7}{8.5}\selectfont\begin{tabular}[t]{@{}c@{}}Spatio-\\temporal\end{tabular}} & {\fontsize{7}{8.5}\selectfont\begin{tabular}[t]{@{}c@{}}All\\tasks\end{tabular}} \\
    \midrule
    \mlogo{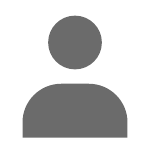}Human & 97.3 & 92.0 & 93.3 & 93.3 & 94.7 & 94.1\std{2.6} & 92.0 & 84.0 & 80.0 & 84.0 & 84.0 & 84.8 \\
    \midrule
    \rowcolor{keybg}
    \multicolumn{13}{c}{\emph{Proprietary MLLMs}} \\
    \mlogo{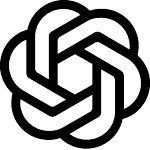}GPT-6 Astra~\citep{openai2026gpt6} & \textbf{72.0} & \textbf{85.3} & \textbf{85.3} & \textbf{76.0} & 52.0 & \textbf{74.1}\std{1.7} & \textbf{52.0} & \textbf{72.0} & \textbf{80.0} & \textbf{64.0} & 24.0 & \textbf{58.4} \\
    \mlogo{openai}GPT-6 Sol~\citep{openai2026gpt6} & 42.7 & 44.0 & 45.3 & 45.3 & 30.7 & 41.6\std{2.4} & 24.0 & 24.0 & 32.0 & 20.0 & \phantom{0}4.0 & 20.8 \\
    \mlogo{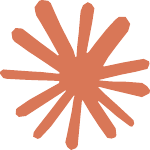}Claude Opus 5.5~\citep{anthropic2026opus55card} & 62.7 & 74.7 & 64.0 & 70.7 & \textbf{54.7} & 65.3\std{0.5} & 40.0 & 48.0 & 48.0 & 60.0 & \textbf{32.0} & 45.6 \\
    \mlogo{claude-color}Claude Fable 5.1~\citep{anthropic2026fable51} & 44.0 & 57.3 & 44.0 & 44.0 & 33.3 & 44.5\std{4.4} & 24.0 & 28.0 & 28.0 & 20.0 & 12.0 & 22.4 \\
    \mlogo{claude-color}Claude Sonnet 5.5~\citep{anthropic2026sonnet55} & 37.3 & 34.7 & 48.0 & 34.7 & 30.7 & 37.1\std{4.7} & 16.0 & \phantom{0}8.0 & 28.0 & 16.0 & \phantom{0}4.0 & 14.4 \\
    \mlogo{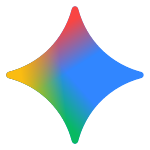}Gemini 4 Argon~\citep{google2026gemini4argon} & 58.7 & 52.0 & 66.7 & 64.0 & 48.0 & 57.9\std{1.2} & 44.0 & 24.0 & 36.0 & 28.0 & 28.0 & 32.0 \\
    \mlogo{gemini-color}Gemini 3.8 Flash~\citep{google2026gemini38flash} & 37.3 & 28.0 & 22.7 & 33.3 & 29.3 & \hlM{30.1\std{1.7}} & 20.0 & \phantom{0}8.0 & \phantom{0}8.0 & 12.0 & 12.0 & \hlR{12.0} \\
    \midrule
    \rowcolor{keybg}
    \multicolumn{13}{c}{\emph{Open-weight MLLMs}} \\
    \mlogo{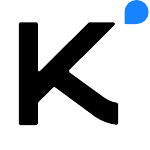}Kimi K3~\citep{kimi2026kimik3} & 37.3 & 33.3 & 21.3 & 36.0 & 32.0 & \hlM{32.0\std{1.4}} & \phantom{0}8.0 & \phantom{0}0.0 & \phantom{0}8.0 & \phantom{0}8.0 & 12.0 & \hlR{\phantom{0}7.2} \\
    \mlogo{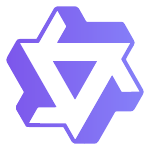}Qwen3.5-9B~\citep{qwen2026qwen35} & \phantom{0}0.0 & \phantom{0}4.0 & \phantom{0}0.0 & \phantom{0}0.0 & \phantom{0}8.0 & \phantom{0}2.4\std{0.0} & \phantom{0}0.0 & \phantom{0}0.0 & \phantom{0}0.0 & \phantom{0}0.0 & \phantom{0}0.0 & \phantom{0}0.0 \\
    \mlogo{qwen-color}Qwen3.8-27B~\citep{qwen2026qwen38} & \phantom{0}6.7 & \phantom{0}4.0 & \phantom{0}5.3 & \phantom{0}4.0 & \phantom{0}9.3 & \phantom{0}5.9\std{0.5} & \phantom{0}0.0 & \phantom{0}0.0 & \phantom{0}0.0 & \phantom{0}0.0 & \phantom{0}0.0 & \phantom{0}0.0 \\
    \mlogo{qwen-color}Qwen3.8-Max~\citep{qwen2026qwen38} & 36.0 & 38.7 & 30.7 & 14.7 & 33.3 & \hlM{30.7\std{3.6}} & \phantom{0}8.0 & 12.0 & \phantom{0}4.0 & \phantom{0}0.0 & 16.0 & \hlR{\phantom{0}8.0} \\
    \midrule
    \rowcolor{keybg}
    \multicolumn{13}{c}{\emph{Open-weight GUI agents}} \\
    \mlogo{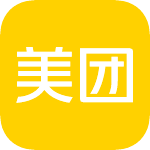}EvoCUA-32B~\citep{xue2026evocua} & \phantom{0}1.3 & \phantom{0}1.3 & \phantom{0}0.0 & \phantom{0}0.0 & 12.0 & \phantom{0}2.9\std{1.2} & \phantom{0}0.0 & \phantom{0}0.0 & \phantom{0}0.0 & \phantom{0}0.0 & \phantom{0}4.0 & \phantom{0}0.8 \\
    \mlogo{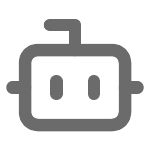}OpenCUA-32B~\citep{wang2025opencua} & \phantom{0}4.0 & \phantom{0}2.7 & \phantom{0}1.3 & \phantom{0}0.0 & \phantom{0}4.0 & \phantom{0}2.4\std{0.8} & \phantom{0}0.0 & \phantom{0}0.0 & \phantom{0}0.0 & \phantom{0}0.0 & \phantom{0}0.0 & \phantom{0}0.0 \\
    \mlogo{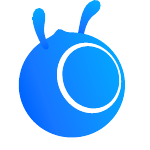}UI-Venus-2-9B~\citep{venus2026uivenus2} & \phantom{0}8.0 & \phantom{0}2.7 & \phantom{0}1.3 & \phantom{0}2.7 & \phantom{0}2.7 & \phantom{0}3.5\std{1.2} & \phantom{0}0.0 & \phantom{0}0.0 & \phantom{0}0.0 & \phantom{0}0.0 & \phantom{0}0.0 & \phantom{0}0.0 \\
    \mlogo{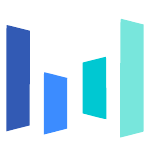}UI-TARS-1.5-7B~\citep{qin2025uitars} & \phantom{0}0.0 & \phantom{0}1.3 & \phantom{0}0.0 & \phantom{0}0.0 & \phantom{0}1.3 & \phantom{0}0.5\std{0.5} & \phantom{0}0.0 & \phantom{0}0.0 & \phantom{0}0.0 & \phantom{0}0.0 & \phantom{0}0.0 & \phantom{0}0.0 \\
    \mlogo{agent}ReCAP-32B-Instruct~\citep{chen2026recap} & \phantom{0}0.0 & \phantom{0}1.3 & \phantom{0}1.3 & \phantom{0}2.7 & \phantom{0}2.7 & \phantom{0}1.6\std{0.8} & \phantom{0}0.0 & \phantom{0}0.0 & \phantom{0}0.0 & \phantom{0}0.0 & \phantom{0}0.0 & \phantom{0}0.0 \\
    \midrule
    \rowcolor{keybg}
    \multicolumn{13}{c}{\emph{Our fine-tuned model}} \\
    \mlogo{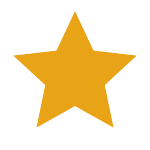}\modelname & 53.3 & 21.3 & 16.0 & 30.7 & 44.0 & \hlM{33.1\std{6.1}} & 20.0 & 12.0 & \phantom{0}4.0 & 12.0 & 16.0 & \hlR{12.8} \\
    \mlogo{qwen-color}Qwen3.5-9B (non-thinking) & \phantom{0}0.0 & \phantom{0}1.3 & \phantom{0}0.0 & \phantom{0}0.0 & \phantom{0}6.7 & \phantom{0}1.6\std{0.0} & \phantom{0}0.0 & \phantom{0}0.0 & \phantom{0}0.0 & \phantom{0}0.0 & \phantom{0}0.0 & \phantom{0}0.0 \\
    \bottomrule
  \end{tabular}}

  \vspace{10pt}
  \captionsetup{type=figure}
  \includegraphics[width=\linewidth]{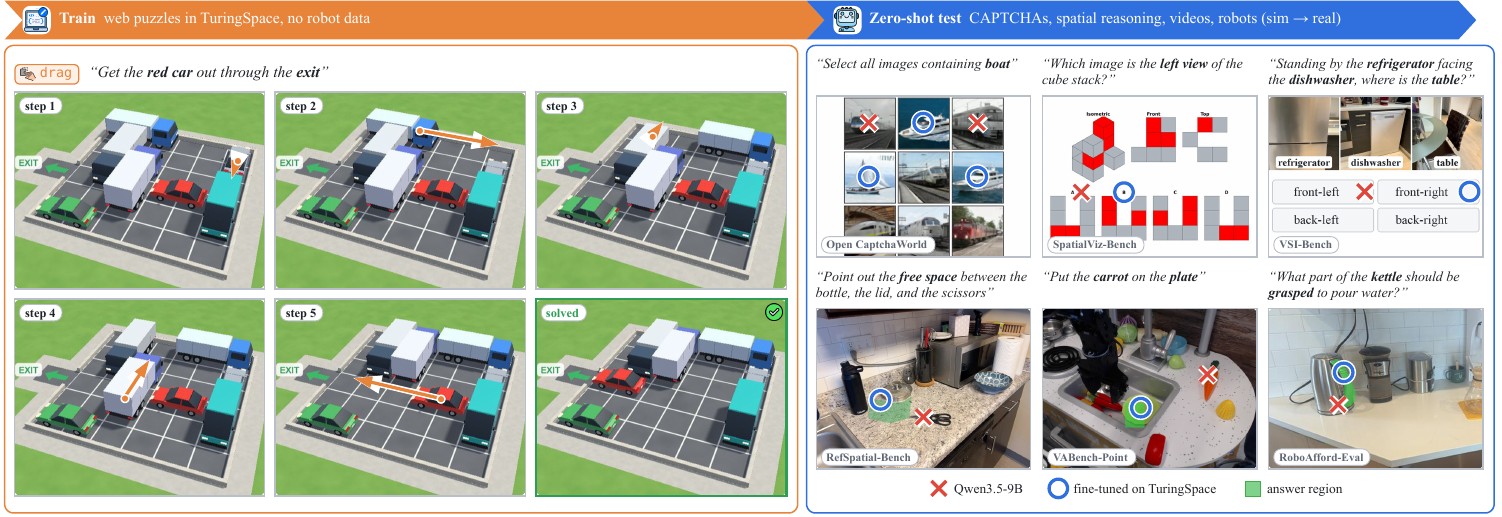}
  \captionof{figure}{From web puzzles to real-world scenes. \textit{Left}: a training trajectory of \gymname{}, in which the model drags the blocking vehicles until the red car can leave; no robot data is used in training. \textit{Right}: zero-shot cases from six other benchmarks, with the answers of Qwen3.5-9B and \modelname{} and the answer region.}
  \label{fig:sft_cases}
\end{table*}

\subsection{Training an MLLM on generated instances}
\label{sec:bench:sft}

To test whether the trajectories of \gymname{} can teach spatial understanding, we fine-tune \texttt{Qwen3.5-9B}~\citep{qwen2026qwen35} on them into \modelname{}.

\noindent\textbf{Training data.}
We run each task in training mode (\cref{eq:splits}) for 500 instances except the spatiotemporal tasks.
For these tasks, we use 200 instances because their trajectories have about 2.3 times as many turns.
This gives 55,000 verified oracle trajectories.
Each action $a^\star_l$ of a trajectory forms one turn with all tool calls issued after observing $o_{l-1}$.

\noindent\textbf{Training objective.}
We train in the thinking-off setting, so each target $a^\star_l$ contains only tool calls, and minimize the cross-entropy loss on these calls,
\begin{equation}
  \mathcal{L}(\theta) = -\frac{1}{Z} \sum_{l=1}^{L} \sum_{t=1}^{|a^\star_l|} \log p_\theta\big(a^\star_{l,t} \,\big|\, h_{l-1},\, a^\star_{l,<t}\big),
  \label{eq:sft}
\end{equation}
where $h_{l-1} = (q, o_0, a^\star_1, \dots, a^\star_{l-1}, o_{l-1})$ is the interaction history, $a^\star_{l,t}$ is the $t$-th token of $a^\star_l$, and $Z = \sum_l |a^\star_l|$ is the number of target tokens.
We update all language-model parameters with \cref{eq:sft} and keep the vision encoder frozen.
\Cref{sec:suppl:sft} gives the full training configuration.

\section{Experiments}
\label{sec:exp}

\subsection{Evaluation protocol}
\label{sec:exp:protocol}

\noindent\textbf{Evaluation setup.}
Each model is evaluated with its official computer-use implementation in OSWorld~\citep{xie2024osworld}, including its system prompt and action format, so that all models are compared fairly.
The system prompt is only adapted to the puzzle page, by removing the sentences about the virtual machine and adding one sentence that authorizes the model to solve CAPTCHAs.
All models receive the same instruction to solve the puzzle and then click the page's ``Verify'' button to submit their answer.
Every model is evaluated on 1920$\times$1080 screenshots with at most 50 steps per attempt, using the sampling parameters of its model card or its API default.
Similarly, Open CaptchaWorld~\citep{luo2025opencaptchaworld} and ReCAP~\citep{chen2026recap} are run with the same setup, so that only the puzzles differ.

We evaluate 16 MLLMs of two kinds.
The first kind is general-purpose MLLMs with a computer-use ability: the proprietary GPT-6 Astra and GPT-6 Sol~\citep{openai2026gpt6}, Claude Opus 5.5~\citep{anthropic2026opus55card}, Claude Fable 5.1~\citep{anthropic2026fable51}, Claude Sonnet 5.5~\citep{anthropic2026sonnet55}, Gemini 4 Argon~\citep{google2026gemini4argon}, and Gemini 3.8 Flash~\citep{google2026gemini38flash}, and the open-weight Kimi K3~\citep{kimi2026kimik3}, Qwen3.5-9B~\citep{qwen2026qwen35}, and Qwen3.8-27B and Qwen3.8-Max~\citep{qwen2026qwen38}.
The second kind is specialized computer-use agents, open-weight models trained to operate GUIs: EvoCUA-32B~\citep{xue2026evocua}, OpenCUA-32B~\citep{wang2025opencua}, UI-Venus-2-9B~\citep{venus2026uivenus2}, UI-TARS-1.5-7B~\citep{qin2025uitars}, and ReCAP-32B-Instruct~\citep{chen2026recap}.
Models with adjustable reasoning effort run at medium, since a human passes a CAPTCHA in about half a minute.
Kimi K3 lacks a medium level and runs at high, and Qwen3.5-9B runs with thinking on and off, as the base of \modelname{}.

\noindent\textbf{Human baseline.}
Three authors who took no part in building or reviewing the tasks each played all 125 tasks on a different test seed, using the same pages and no time limit. Only the first submission counts.

\noindent\textbf{Evaluation metrics.}
Each task has three test instances, and a model plays each once.
An instance is solved only if the checker accepts the state that the model submits.
Unless stated otherwise, pass@1 is the success rate averaged over the three seeds (375 attempts) with its standard deviation, and \passk{} is the fraction of tasks solved in all three instances, which measures stability.

\subsection{How well do frontier MLLMs perform?}
\label{sec:exp:main}

\noindent\textbf{Frontier MLLMs fall well short of humans and solve few tasks reliably.}
\Cref{tab:main} shows a large gap between humans and every model.
Humans solve 94.1\% of the tasks and need only 33.1\,s per task on average, whereas the strongest model, \mbox{GPT-6} Astra, reaches 74.1\% pass@1.
Only Claude Opus 5.5 and Gemini 4 Argon also exceed 50\%, and the other proprietary models fail more than half of the tasks.
The gap widens when a task must be solved on all three instances: \passk{} drops to 58.4\% for \mbox{GPT-6} Astra and lower for the other models, against 84.8\% for humans.
This drop shows that models often solve a task on one instance but fail it on another.
Even the strongest models therefore lack the reliable spatial understanding that humans show.

\noindent\textbf{Spatiotemporal tasks are the hardest category for the strongest models.}
For almost all proprietary models, the spatiotemporal tasks are the hardest of the five categories.
\mbox{GPT-6} Astra, for example, reaches only 52.0\% on the spatiotemporal tasks.
The seven proprietary models also score closer to one another here than on any other category.
This is also the only category in which \mbox{GPT-6} Astra is not the best model, as Claude Opus 5.5 scores higher.
Following how a scene changes over time thus remains challenging for all models, and the strongest ones gain little on it.

\noindent\textbf{Open-weight models are far behind proprietary models, but training in \gymname{} narrows the gap.}
The best open-weight MLLMs, Kimi K3 and Qwen3.8-Max, have 2.8 and 2.4 trillion parameters~\citep{kimi2026kimik3,qwen2026qwen38}, yet both fall below every proprietary model except Gemini 3.8 Flash.
GUI agents solve almost none of the tasks, although UI-Venus-2-9B and ReCAP-32B-Instruct are trained to solve common CAPTCHAs~\citep{venus2026uivenus2,chen2026recap}.
In contrast, fine-tuning \texttt{Qwen3.5-9B} without thinking in \gymname{} raises its pass@1 from 1.6\% to 33.1\%.
The resulting \modelname{} outperforms Kimi K3, Qwen3.8-Max, and Gemini 3.8 Flash in both metrics and comes close to Claude Sonnet 5.5.
Training in \gymname{} thus narrows the gap to proprietary models, letting a 9B model surpass open-weight MLLMs over 250 times its size.

\subsection{Transfer beyond \benchname{}}
\label{sec:exp:sft}

We test the transfer of \modelname{} in two directions (\cref{tab:sft}).
Within CAPTCHAs, we select ReCAP~\citep{chen2026recap} and Open CaptchaWorld~\citep{luo2025opencaptchaworld}, which contain 7 and 19 types of conventional CAPTCHAs.
Beyond CAPTCHAs, we select benchmarks that mainly require spatial understanding, so any gain reflects what \bench{} teaches.
SpatialViz-Bench~\citep{wang2026spatialviz} evaluates mental rotation, folding, and projection through multiple-choice questions, and VSI-Bench~\citep{yang2025thinkinginspace} asks about the directions, distances, and sizes of objects in videos.
Among the three robot benchmarks, RefSpatial-Bench~\citep{zhou2025roborefer} requires locating objects and free space by spatial relations, VABench-Point~\citep{yuan2026fsd}{\parfillskip=0pt\par}
\begin{wraptable}{r}{0.45\linewidth}
  \vspace{-1.7pt} %
  \centering
  \scriptsize %
  \setlength{\tabcolsep}{2pt}
  \renewcommand{\arraystretch}{1.2}
  \caption{Success rate (\%) of \texttt{Qwen3.5-9B} (Base) and \modelname{} (SFT) on \benchname{} and seven out-of-distribution benchmarks.}
  \label{tab:sft}
  \begin{tabular}{@{}l>{\scriptsize\color{black!65}}lccc@{}}
    \toprule
    Benchmark & Tests & Base & SFT & $\Delta$ \\
    \midrule
    \rowcolor{keybg}
    \multicolumn{5}{@{}l}{\emph{In-domain}} \\
    \benchname{}       & spatial CAPTCHAs & \phantom{0}1.6 & 33.1 & +31.5 \\
    \rowcolor{keybg}
    \multicolumn{5}{@{}l}{\emph{CAPTCHA}} \\
    ReCAP              & 7 types  & 56.4 & 73.6 & +17.2 \\
    Open CaptchaWorld  & 19 types & 26.9 & 36.9 & +10.0 \\
    \rowcolor{keybg}
    \multicolumn{5}{@{}l}{\emph{Spatial visualization}} \\
    SpatialViz-Bench   & visualization & 28.1 & 34.2 & +6.1 \\
    \rowcolor{keybg}
    \multicolumn{5}{@{}l}{\emph{Spatial reasoning in videos}} \\
    VSI-Bench          & indoor scenes & 50.5 & 60.4 & +9.9 \\
    \rowcolor{keybg}
    \multicolumn{5}{@{}l}{\emph{Robot manipulation in real scenes}} \\
    RefSpatial-Bench   & spatial referring  & 38.3 & 58.1 & +19.8 \\
    VABench-Point      & where to act       & 35.7 & 43.1 & +7.4 \\
    RoboAfford-Eval    & affordances & 67.6 & 79.8 & +12.2 \\
    \bottomrule
    \end{tabular}
  \vspace{-0.6em}
\end{wraptable}
\noindent requires predicting where a robot should act, and RoboAfford-Eval~\citep{hao2025roboafford} requires identifying the part of an object that affords a given action.
We evaluate the base model, \texttt{Qwen3.5-9B} without thinking, and \modelname{} zero-shot with the official prompts of each benchmark.

\noindent\textbf{Spatial CAPTCHAs teach conventional CAPTCHAs, but not the reverse.}
\modelname{} improves on ReCAP and Open CaptchaWorld, although none of their puzzle types appears in \bench{}.
Conversely, agents trained on such CAPTCHAs barely solve \bench{} (\cref{sec:exp:main}).
Learning spatial CAPTCHAs therefore teaches conventional ones, whereas learning conventional CAPTCHAs does not teach spatial ones.

\noindent\textbf{Training on synthetic web puzzles generalizes to real-world scenes.}
We next test benchmarks whose inputs differ from our rendered web pages.
\modelname{} improves on the multiple-choice questions of SpatialViz-Bench and the videos of real rooms in VSI-Bench, so the spatial skills learned by acting do not depend on pointing at the screen.
It also improves on all three robot benchmarks, whose photos of real robot workspaces never appear in our training data.
In the robot cases of \cref{fig:sft_cases}, the base model points at an object that the instruction names, such as the scissors or the carrot, whereas \modelname{} points where the action takes effect: the free space, the plate, or the handle to grasp.
Our tasks train this decision.
In the training trajectory on the left of \cref{fig:sft_cases}, the model must decide where each blocking vehicle has to go so that the red car can leave, and it sees the result of every drag before the next one.
The loop of acting, observing, and adjusting thus generalizes to real-world scenes.

\section{Analysis}
\label{sec:analysis}

\subsection{Does more thinking lead to more success?}
\label{sec:analysis:thinking}

\begin{figure*}[!t]
  \centering
  \includegraphics[width=\linewidth]{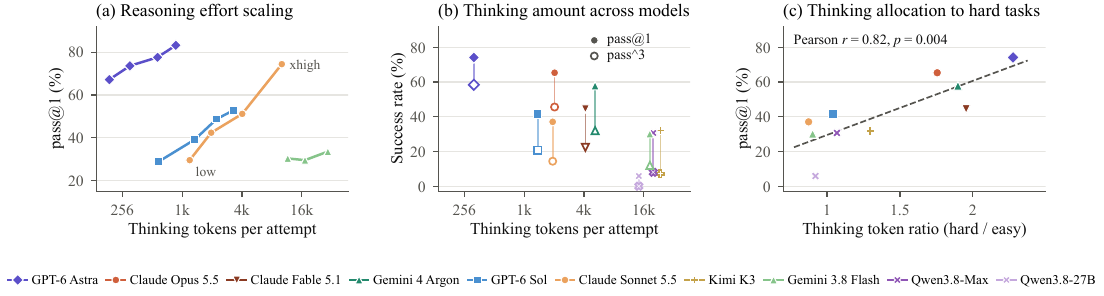}
  \captionof{figure}{Thinking tokens and success rate: (a) raising the reasoning effort of four models, (b) comparing the ten models at medium effort, and (c) the ratio of thinking on the hardest to the easiest quartile of tasks, against pass@1.}
  \label{fig:thinking}

  \vspace{3pt}
  \includegraphics[width=\linewidth]{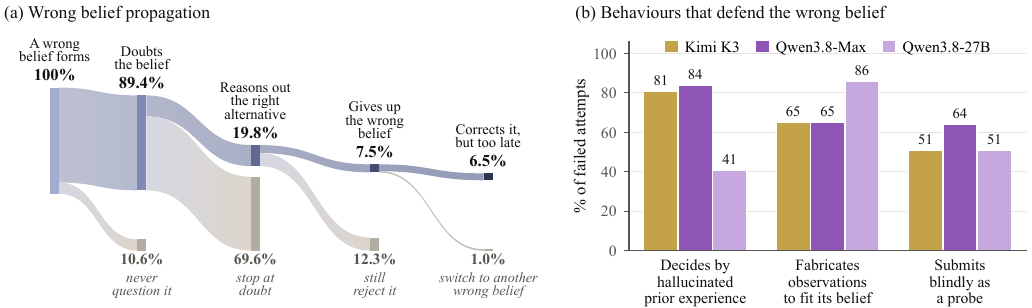}
  \captionof{figure}{Wrong beliefs in the 293 failed attempts of Kimi K3, Qwen3.8-Max, and Qwen3.8-27B. (a) Share of attempts that reach each level of revising the decisive wrong belief; the grey part of a bar stops at the level above. (b) Share of attempts showing each behaviour.}
  \label{fig:belief}
\end{figure*}

Frontier MLLMs can think for thousands of tokens before each action, and most let the user set this budget through a reasoning effort.
We ask whether thinking longer helps a model solve our tasks (\cref{fig:thinking}).

\noindent\textbf{Raising the reasoning effort helps only some models.}
We first rerun all 125 tasks at seed 36 for four models, at every reasoning effort from the lowest to the highest level. As shown in \cref{fig:thinking}a, Claude Sonnet 5.5 rises from 29.6\% to 74.4\%, \mbox{GPT-6} Sol from 28.8\% to 52.8\%, and \mbox{GPT-6} Astra from 67.2\% to 83.2\%.
Gemini 3.8 Flash, in contrast, stays between 29.6\% and 33.6\% at all levels.
Therefore, more thinking helps some models, but this does not show that thinking is what makes a model strong.

\noindent\textbf{Stronger MLLMs do not think more, but know where thinking matters.}
We next ask whether the amount of thinking explains why some models are stronger.
Across the ten models at medium effort, the log of a model's mean thinking tokens per attempt correlates negatively with its pass@1 (Pearson $r=-0.75$, $p=0.01$) and with its \passk{} ($r=-0.82$, $p=0.004$; \cref{fig:thinking}b).
GPT-6 Astra, the strongest model, uses only 319 thinking tokens per attempt, while Kimi K3 uses 24,015 and reaches 32.0\%.
Hence, the amount of thinking does not separate the models, but stronger models may know how to allocate thinking better.

To measure this, we rank the tasks by the mean failure rate of the ten models and divide each model's mean thinking tokens on the hardest quartile by that on the easiest quartile (\cref{fig:thinking}c).
GPT-6 Astra thinks 2.3 times as much on the hardest quartile, Claude Fable 5.1 2.0 times, Gemini 4 Argon 1.9 times, and Claude Opus 5.5 1.8 times, whereas Claude Sonnet 5.5, Gemini 3.8 Flash, and Qwen3.8-Max think at most 1.07 times as much.
This ratio correlates with pass@1 (Pearson $r=0.82$, $p=0.004$) and with \passk{} ($r=0.83$, $p=0.003$).
Stronger models therefore succeed not by thinking more but by knowing where to think, whereas weaker models think almost equally long on easy and hard tasks, so much of their reasoning goes to tasks that do not need it.
This leaves a question: \emph{when models think at length, what does their long reasoning do?}

\subsection{Can reasoning correct a wrong belief?}
\label{sec:analysis:belief}

To answer this question, we collect all 293 failed attempts at seed 36 of Kimi K3, Qwen3.8-Max, and Qwen3.8-27B.
They are among the models that think the most, and because they are open-weight, we can access their full reasoning.
We first read a sample of these traces by hand and find two patterns.
First, each failure comes from one wrong belief, for example a wrong reading of the screen.
We call it the decisive wrong belief, and \cref{fig:belief}a shows how far the model gets in correcting it.
Second, the model often shows three behaviours that keep this belief in place (\cref{fig:belief}b).
Claude Opus 5.5 then labels every attempt with a prompt built from these patterns.
One author checked the labels of 50 sampled attempts and agreed with 92\% of them, and \cref{sec:suppl:belief,sec:suppl:prompt} give representative examples and the prompt, respectively.

\noindent\textbf{A wrong belief forms in the first steps and propagates through the later reasoning to the final answer.}
We first ask when the wrong belief forms and whether later reasoning changes it.
In 68\% of the attempts, the belief appears within the first two steps, when the median attempt has written only 18\% of its thinking tokens.
This means that more than four fifths of the reasoning builds on a wrong belief.
This reasoning questions the belief at least once in 89.4\% of the attempts, yet only 7.5\% give up the wrong belief by the end.
In almost all of the 6.5\% of attempts that switch to the correct belief, the model sees its error only when the screen shows that the puzzle can no longer be solved, for example after a wrong move has ended the game.
As a result, the wrong belief from the first steps decides the outcome, and later reasoning keeps reinforcing this wrong belief.

\noindent\textbf{Even when a model reasons out the correct solution, it usually does not adopt it.}
We next ask how often a model reasons out the correct solution.
Here, reasoning out the correct solution means that the reasoning reaches the correct answer or most of the solution and supports it with an observation or a computation.
As shown in \cref{fig:belief}a, only 19.8\% of the failed attempts reason out the correct solution, while the other doubting attempts only question the belief, name the correct alternative as an untested possibility, or work out part of it.
However, most of these attempts still keep the wrong belief: in 36 of the 58 attempts that reason out the correct solution, the model returns to its first, wrong belief.
One straightforward example is a puzzle that asks where a falling block will land: the model computes the landing cell and states it as its conclusion, but then submits the cell it chose first because it is unsure of the fall height.
The failure therefore lies not only in finding a new correct solution path but also in accepting it.

\noindent\textbf{Models tend to hallucinate evidence to defend their wrong belief, especially when they reason out a correct alternative or when they think longer.}
We next ask what the later reasoning does with the wrong belief.
\cref{fig:belief}b shows, for each model, three behaviours that keep the wrong belief in place.
Over all 293 failed attempts, the model fabricates an observation in 74\% of them: it reports a detail that the screenshot does not show, so that the screenshot description seems to fit its belief.
In 66\% of them, the model decides by a hallucinated prior experience: it assumes how such puzzles usually work and imagines a potential solution path based on this bias, although the screen never says so.
This suggests that these models may rely on their world knowledge to guess a plausible path or to justify their wrong belief, instead of carefully examining what the screenshot actually shows.
In 55\% of them, the model submits blindly as a probe: it clicks Verify while unsure, expecting extra feedback and a second chance.

These behaviours are especially frequent when the model has a strong reason to change its mind.
Among the 36 attempts that reason out the correct solution but still reject it, 86\% fabricate an observation, 89\% decide by a hallucinated prior experience, and 81\% submit blindly.
Among the 31 attempts that never question the belief, the same three shares are only 55\%, 16\%, and 32\%.
They also grow with reasoning length: from the shortest to the longest quartile of thinking tokens within each model, fabricated observations rise from 58\% to 85\%, hallucinated prior experience from 41\% to 85\%, and blind submissions from 36\% to 74\%.
Taken together, long reasoning defends the wrong belief instead of checking it against the screenshot.
This explains why thinking longer alone does not bring success (\cref{sec:analysis:thinking}).

\subsection{Which failures do frontier MLLMs share?}
\label{sec:analysis:failures}

Finally, we ask which failures remain in the three strongest models: GPT-6 Astra, Claude Opus 5.5, and Gemini 4 Argon.
We label 380 of their 385 failed attempts with one of ten root causes in three groups: perception, reasoning, and interaction.
We discard the other five because Gemini 4 Argon repeatedly refuses to answer them. Perception errors are the most common and decide 57.1\% of the failures. Visual feedback does not remove these errors: closed-loop manipulation and planning tasks show a new screenshot after every action, yet perception errors still cause 85 of their 129 failures.
In contrast, no failure comes from a lack of grounding ability, such as clicking at the wrong coordinates. Spatial perception is therefore the main bottleneck of frontier MLLMs on our tasks. \Cref{sec:suppl:errors} shows the full analysis.
\section{Conclusion}
\label{sec:conclusion}

We introduce \bench{}, a synthetic web environment of 125 interactive spatial CAPTCHAs that MLLMs must solve through a loop of acting. On \benchname{}, frontier MLLMs fall well short of humans, reaching at best 74.1\% against 94.1\%. Fine-tuning \texttt{Qwen3.5-9B} in \gymname{} raises its success rate from 1.6\% to 33.1\% and improves it on all seven OOD benchmarks. Spatial intelligence can thus be learned through interactive synthetic web puzzles and transfer to real-world scenes.

\bibliographystyle{assets/plainnat}
\bibliography{related}

\clearpage
\newpage
\beginappendix
\vspace{0.5em}
\newcommand{\suppentry}[3]{\noindent\hyperref[#1]{\textbf{\ref*{#1}\quad #2}}\ \dotfill\ \pageref*{#1}\par
  \noindent\hspace{1.45em}{\small\color{black!62}#3}\par\medskip}
\suppentry{sec:suppl:sft}{Training details}{Action format and full training configuration of \modelname (\cref{sec:bench:sft}).}
\suppentry{sec:suppl:ood}{Out-of-domain evaluation}{Benchmarks, settings, and prompts of the transfer evaluation (\cref{sec:exp:sft}).}
\suppentry{sec:suppl:belief}{Wrong belief propagation}{Annotation, levels, and behaviours of the decisive wrong belief, with nine annotated cases (\cref{sec:analysis:belief}).}
\suppentry{sec:suppl:errors}{Error analysis}{Labelling procedure, failure taxonomy, and findings for the three strongest models (\cref{sec:analysis:failures}).}
\suppentry{sec:suppl:compare}{Comparison with related benchmarks}{\bench compared with CAPTCHA benchmarks and interactive spatial benchmarks.}
\suppentry{sec:suppl:eval_prompts}{Evaluation prompts}{System prompt and first user message of every evaluated model (\cref{sec:exp:protocol}).}
\suppentry{sec:suppl:prompt}{Annotation prompts}{Prompts of the agent-as-a-judge for the wrong-belief and root-cause analyses (\cref{sec:analysis:belief,sec:analysis:failures}).}
\suppentry{sec:suppl:ethics}{Ethics statement}{Purpose and dual use of \bench.}
\newenvironment{suppblock}{\par\smallskip\noindent\begin{minipage}{\linewidth}\centering}{\end{minipage}\par\smallskip}
\lstdefinestyle{evalprompt}{basicstyle=\fontencoding{T1}\fontfamily{lmtt}\selectfont\fontsize{6.6}{7.9}\selectfont, columns=fixed, basewidth=0.525em,
  breaklines=true, breakatwhitespace=true, showstringspaces=false, upquote=true, frame=none, xleftmargin=0pt,
  aboveskip=1pt, belowskip=4pt, breakindent=10pt, breakautoindent=true, %
  moredelim=[is][\color{keyblue}]{@@}{@@}, moredelim=[is][\color{black!50}\itshape]{!!}{!!}}
\newcommand{\promptpart}[2]{\par\noindent{\color{keyblue}\sffamily\bfseries\scriptsize #1}\par\lstinputlisting[style=evalprompt]{sec/prompts/#2.txt}}
\newtcolorbox{promptbox}[2][]{enhanced, colback=keybg, colframe=keyblue, boxrule=0.6pt, arc=2pt,
  colbacktitle=keyblue, coltitle=white, fonttitle=\bfseries\small, toptitle=2.5pt, bottomtitle=2.5pt,
  left=7pt, right=7pt, top=5pt, bottom=3pt, before skip=10pt, after skip=0pt, title={#2}, #1}
\ifdefined\nolinenumbers
  \BeforeBeginEnvironment{promptbox}{\nolinenumbers}
  \AfterEndEnvironment{promptbox}{\linenumbers}
\fi
\clearpage

\section{Training details}
\label{sec:suppl:sft}

This section gives the data format and the full configuration of the fine-tuning in \cref{sec:bench:sft}.

\medskip\noindent\textbf{Action format.}
The model calls one computer-use tool with 19 mouse and keyboard actions, which is the tool that the Qwen agent uses in evaluation.
Coordinates lie on a $1000 \times 1000$ grid, and a pixel $x$ of a screenshot of width $W$ maps to $\mathrm{round}(999\, x / (W - 1))$, and likewise for $y$.
Each sample registers the tool definition and has no system prompt.
In the thinking-off setting, every turn starts with an empty thinking block, which carries no loss.
A drag is one turn that moves the cursor to the start and then drags to the end.
A held drag spans several turns: the first turn moves the cursor to the control, presses the button, and makes the first move, each later turn makes one further move after a new screenshot, and the last turn releases the button.
Every trajectory has at most 18 turns, so each trajectory forms one sample that keeps all of its screenshots.

\medskip\noindent\textbf{Training sample.}
\Cref{fig:suppl:train_system,fig:suppl:train_sample} show the first turn of a training sample.
Its system turn is the tool definition alone, as the chat template of Qwen3.5 renders it from the registered tool.
The first user turn holds the first screenshot and the instruction of the evaluation.
Each later turn holds the tool calls of the model and one tool response with the next screenshot.
\modelname is evaluated with the same messages, so each turn in evaluation is identical to a training turn.

\medskip\noindent\textbf{Screen layout.}
Each trajectory shows the puzzle card at a random position on the screen.
The card is enlarged by a random factor between 1.0 and 1.2 and keeps at least 12 pixels from every screen edge.
The page background takes one of ten colours.
Every coordinate of the oracle actions is mapped onto the moved card, and a trajectory is kept only if the page judges its replay solved.
Benchmark instances keep the default layout.

\medskip\noindent\textbf{Configuration.}
\Cref{tab:suppl:config} lists the training configuration.

\begin{suppblock}
  \captionsetup{type=table} %
  \small
  \setlength{\tabcolsep}{4pt}
  \captionof{table}{Training configuration of \modelname.}
  \label{tab:suppl:config}
  \begin{tabular*}{\linewidth}{@{\extracolsep{\fill}}llll@{}}
    \toprule
    Configuration & Value & Configuration & Value \\
    \midrule
    Base model & \texttt{Qwen3.5-9B}, thinking off & Global batch size & 128 samples \\
    Trainable parameters & language model (full) & Micro batch size & 1 per GPU \\
    Frozen parameters & vision encoder and aligner & Gradient accumulation & 4 steps \\
    Framework & ms-swift, Megatron backend & Training steps & 1,000 (about 2.3 epochs) \\
    Hardware & 32$\times$ NVIDIA H200 & Optimizer & Adam \\
    Parallelism & tensor 1, data 32 & Adam $(\beta_1, \beta_2, \epsilon)$ & $(0.9,\ 0.95,\ 10^{-8})$ \\
    Training trajectories & 55,000 & Weight decay & 0.033 \\
    Validation trajectories & 375 (3 per task) & Learning rate & $5 \times 10^{-6}$, cosine decay to $10^{-6}$ \\
    Captured screenshots & $1920 \times 1080$ & Warmup & 3\% of steps (30 steps) \\
    Training resolution & $1280 \times 720$ & Gradient clipping & 1.0 \\
    Tokens per screenshot & 880 & Precision & bf16, fp32 optimizer states \\
    Screenshots per sample & at most 18 & Recomputation & full \\
    Maximum sequence length & 20,480 tokens & Loss & \cref{eq:sft}, token average \\
    Sequence packing & off & Random seed & 1234 \\
    \bottomrule
  \end{tabular*}
\end{suppblock}

\par\begingroup\captionsetup{type=figure}%
\begin{promptbox}{System turn of every training sample}
  \promptpart{Rendered from the registered computer-use tool}{train_system}
\end{promptbox}\nopagebreak\vspace{9pt}\nopagebreak
\captionof{figure}{System turn of every training sample.}
\label{fig:suppl:train_system}
\par\endgroup\medskip

\par\begingroup\captionsetup{type=figure}%
\begin{promptbox}{First turn of a training sample of \texttt{backmost\_layer\_select}}
  \promptpart{First user turn}{train_user}
  \promptpart{Assistant turn (target)}{train_turn}
  \promptpart{Tool response}{train_tool}
\end{promptbox}\nopagebreak\vspace{9pt}\nopagebreak
\captionof{figure}{First turn of a training sample of \texttt{backmost\_layer\_select}.}
\label{fig:suppl:train_sample}
\par\endgroup\medskip

\clearpage
\section{Out-of-domain evaluation}
\label{sec:suppl:ood}

This section gives the benchmarks, settings, and prompts of the transfer evaluation in \cref{sec:exp:sft}.

\subsection{Benchmarks}
\label{sec:suppl:ood:benchmarks}

\Cref{tab:suppl:ood_bench} summarizes the seven benchmarks.

\begin{suppblock}
\captionsetup{type=table} %
\small
\setlength{\tabcolsep}{4pt}
\captionof{table}{The seven out-of-domain benchmarks.}
\label{tab:suppl:ood_bench}
\begin{tabular*}{\linewidth}{@{\extracolsep{\fill}}llll>{\raggedright\arraybackslash}p{2.6cm}@{}}
\toprule
Benchmark & Input & Size & Answer & Score \\
\midrule
ReCAP & CAPTCHA web page & 140 puzzles, 7 types & GUI actions & solved puzzles \\
Open CaptchaWorld & CAPTCHA web page & 453 puzzles, 19 types & GUI actions & solved puzzles \\
SpatialViz-Bench & one image & 1,180 questions, 12 tasks & option letter & accuracy \\
VSI-Bench & 32 video frames & 5,130 questions, 8 tasks & option letter or number & accuracy or MRA, mean over tasks \\
RefSpatial-Bench & one photo & 277 questions, 3 splits & points & share of points in the mask \\
VABench-Point & one robot photo & 300 questions & points & share of points in the mask \\
RoboAfford-Eval & one tabletop photo & 338 questions & points & share of points in the mask \\
\bottomrule
\end{tabular*}
\end{suppblock}

\medskip\noindent\textbf{ReCAP.}
ReCAP~\citep{chen2026recap} serves seven types of conventional CAPTCHAs on web pages, namely text, compact text, icon selection, paged, icon match, slider, and image grid.
We generate 20 puzzles of each type.

\medskip\noindent\textbf{Open CaptchaWorld.}
Open CaptchaWorld~\citep{luo2025opencaptchaworld} contains interactive web CAPTCHAs such as dice counting, path finding, and rotation matching.
We run all puzzles of its public repository except the 10 puzzles of Hold\_Button, which gives 453 puzzles of 19 types.
We reword the question of Dice\_Count from ``Sum up the numbers on all the dice'' to ``Sum up the numbers on top of all the dice''.
We also reword the question of Unusual\_Detection from ``Select the animal with the wrong head'' to ``Select all the animals with the wrong head''.
Each page shows a single puzzle, and the server of Open CaptchaWorld judges the first answer that the page submits.

\medskip\noindent\textbf{SpatialViz-Bench.}
SpatialViz-Bench~\citep{wang2026spatialviz} tests spatial visualization with 1,180 multiple-choice questions.
Its 12 tasks cover mental rotation, mental folding, visual penetration, and mental animation.
Each question shows one image with up to four options, and the score is the accuracy of the chosen letter.

\medskip\noindent\textbf{VSI-Bench.}
VSI-Bench~\citep{yang2025thinkinginspace} asks 5,130 questions about videos of real indoor scenes.
Its eight tasks are object counting, absolute distance, object size, room size, relative distance, relative direction, route planning, and appearance order.
Each video enters as 32 uniformly sampled frames at their original resolution.
Multiple-choice questions are scored by accuracy and numerical questions by mean relative accuracy (MRA) over the thresholds from 0.50 to 0.95.
The score averages the eight tasks after the three difficulty levels of relative direction are averaged into one.

\medskip\noindent\textbf{RefSpatial-Bench.}
RefSpatial-Bench~\citep{zhou2025roborefer} asks a model to point at an object or a free region that a spatial relation describes in a real photo.
Its 277 questions fall into the Location, Placement, and Unseen splits.
The score of a question is the share of predicted points that fall inside its target mask.

\medskip\noindent\textbf{VABench-Point.}
VABench-Point~\citep{yuan2026fsd} asks a model to point where a robot should act to complete a task instruction.
Its 300 photos come from the robot datasets Bridge, DROID, and RT-X.
The score of a question is the share of predicted points inside its target mask.

\medskip\noindent\textbf{RoboAfford-Eval.}
RoboAfford-Eval~\citep{hao2025roboafford} contains 338 questions on tabletop photos.
Each question asks for an object, the part of an object to grasp, or a free region to place an item.
The score of a question is the share of predicted points inside its target mask.

\subsection{Evaluation settings}
\label{sec:suppl:ood:settings}

Both models run zero-shot with thinking off.

\medskip\noindent\textbf{CAPTCHA benchmarks.}
ReCAP and Open CaptchaWorld run in the harness of \bench{} with the settings of \cref{sec:exp:protocol}.
\texttt{Qwen3.5-9B} uses the Qwen agent with the sampling parameters of its model card.
\modelname uses the messages of its training samples with greedy decoding.
Each benchmark replaces the instruction of \bench{} with its own task instruction in \cref{fig:suppl:ood_captcha}.

\medskip\noindent\textbf{Single-turn benchmarks.}
SpatialViz-Bench, VSI-Bench, and RefSpatial-Bench use the prompts and answer parsers of VLMEvalKit.
VABench-Point and RoboAfford-Eval use those of EmbodiedEvalKit.
Both models receive the same prompts.
\Cref{tab:suppl:ood_settings} lists the input and decoding of each model.
On SpatialViz-Bench and RefSpatial-Bench, \texttt{Qwen3.5-9B} samples with the Qwen3.5 configuration of VLMEvalKit.
On VABench-Point and RoboAfford-Eval, it samples with the default configuration of EmbodiedEvalKit.

\begin{suppblock}
\captionsetup{type=table} %
\small
\setlength{\tabcolsep}{5pt}
\captionof{table}{Input and decoding of the five single-turn benchmarks.}
\label{tab:suppl:ood_settings}
\begin{tabular*}{\linewidth}{@{\extracolsep{\fill}}>{\raggedright\arraybackslash}p{2.9cm}l>{\raggedright\arraybackslash}p{2.5cm}>{\raggedright\arraybackslash}p{4.6cm}r@{}}
\toprule
Benchmark & Model & Input & Decoding & Max.\ new tokens \\
\midrule
SpatialViz-Bench, RefSpatial-Bench & \texttt{Qwen3.5-9B} & original image & temperature 1.0, top-$p$ 0.95, top-$k$ 20, presence penalty 1.5 & 32,768 \\
 & \modelname & $1280 \times 720$ & greedy & 256 \\
\midrule
VSI-Bench & \texttt{Qwen3.5-9B} & 32 original frames & greedy & 256 \\
 & \modelname & 32 original frames & greedy & 256 \\
\midrule
VABench-Point, RoboAfford-Eval & \texttt{Qwen3.5-9B} & original image & temperature 0.7, top-$p$ 0.8, top-$k$ 20, repetition penalty 1.05 & 4,096 \\
 & \modelname & $1280 \times 720$ & greedy & 256 \\
\bottomrule
\end{tabular*}
\end{suppblock}

\subsection{Prompts}
\label{sec:suppl:ood:prompts}

\Cref{fig:suppl:ood_captcha,fig:suppl:ood_spatialviz,fig:suppl:ood_vsi,fig:suppl:ood_ref,fig:suppl:ood_robot} give the prompts.
In the boxes, {\color{black!50}\fontfamily{lmtt}\selectfont\itshape [image]} marks an image and other bracketed grey text marks a field of the item, such as the question or its options.

\par\begingroup\captionsetup{type=figure}%
\begin{promptbox}{Task instructions of ReCAP and Open CaptchaWorld}
  \promptpart{ReCAP}{ood_recap}
  \promptpart{Open CaptchaWorld}{ood_ocw}
\end{promptbox}\nopagebreak\vspace{9pt}\nopagebreak
\captionof{figure}{Task instructions of ReCAP and Open CaptchaWorld.}
\label{fig:suppl:ood_captcha}
\par\endgroup\medskip

\par\begingroup\captionsetup{type=figure}%
\begin{promptbox}{Prompt of SpatialViz-Bench}
  \promptpart{User message}{ood_spatialviz}
\end{promptbox}\nopagebreak\vspace{9pt}\nopagebreak
\captionof{figure}{User message of SpatialViz-Bench.}
\label{fig:suppl:ood_spatialviz}
\par\endgroup\medskip

\par\begingroup\captionsetup{type=figure}%
\begin{promptbox}{Prompts of VSI-Bench}
  \promptpart{User message of a multiple-choice question}{ood_vsi_mcq}
  \promptpart{User message of a numerical question}{ood_vsi_na}
\end{promptbox}\nopagebreak\vspace{9pt}\nopagebreak
\captionof{figure}{User messages of VSI-Bench.}
\label{fig:suppl:ood_vsi}
\par\endgroup\medskip

\par\begingroup\captionsetup{type=figure}%
\begin{promptbox}{Prompt of RefSpatial-Bench}
  \promptpart{User message}{ood_refspatial}
\end{promptbox}\nopagebreak\vspace{9pt}\nopagebreak
\captionof{figure}{User message of RefSpatial-Bench.}
\label{fig:suppl:ood_ref}
\par\endgroup\medskip

\par\begingroup\captionsetup{type=figure}%
\begin{promptbox}{Prompts of VABench-Point and RoboAfford-Eval}
  \promptpart{System prompt of both benchmarks}{ood_eek_system}
  \promptpart{User message of VABench-Point}{ood_vabench}
  \promptpart{User message of RoboAfford-Eval}{ood_roboafford}
\end{promptbox}\nopagebreak\vspace{9pt}\nopagebreak
\captionof{figure}{System prompt and user messages of VABench-Point and RoboAfford-Eval.}
\label{fig:suppl:ood_robot}
\par\endgroup\medskip

\clearpage
\section{Wrong belief propagation}
\label{sec:suppl:belief}

\definecolor{casewrong}{HTML}{C62828}
\definecolor{caseright}{HTML}{2E7D32}
\newtcolorbox{casebox}[1]{enhanced, colback=white, colframe=keyblue, boxrule=0.6pt, arc=2pt, left=7pt, right=7pt, top=8pt, bottom=4pt,
  title={#1}, fonttitle=\bfseries\small, coltitle=white, colbacktitle=keyblue,
  attach boxed title to top left={xshift=8pt, yshift*=-\tcboxedtitleheight/2}, boxed title style={arc=1.5pt, boxrule=0pt, left=4pt, right=4pt, top=2pt, bottom=2pt},
  fontupper=\small, before skip=10pt, after skip=6pt}
\newtcolorbox{casepanel}[1]{enhanced, frame hidden, colback=#1!6, borderline west={2pt}{0pt}{#1}, sharp corners, left=6pt, right=5pt, top=2.5pt, bottom=2.5pt, boxsep=0pt, before skip=0pt, after skip=4pt, fontupper=\small}
\newcommand{\casetask}[1]{\noindent{\bfseries Task.}\ #1\par\vspace{4pt}}
\newcommand{\casebelief}[1]{\begin{casepanel}{casewrong}\textcolor{casewrong}{\bfseries Wrong belief.}\ #1\end{casepanel}}
\newcommand{\casetruth}[1]{\begin{casepanel}{caseright}\textcolor{caseright}{\bfseries In fact.}\ #1\end{casepanel}}
\newcommand{\casereasoning}{\par\vspace{5pt}\noindent{\bfseries Reasoning} {\color{black!55}\footnotesize(verbatim excerpts)}\par\vspace{1pt}}
\newcommand{\casestep}[1]{\textcolor{black!55}{\textsf{\scriptsize Step~#1}}}
\newcommand{\casenote}[1]{\par{\color{black!62}\itshape\footnotesize$\hookrightarrow$~#1}}
\newcommand{\caseoutcome}[1]{\par\vspace{4pt}\begin{tcolorbox}[enhanced, frame hidden, colback=black!4, arc=1.5pt, left=6pt, right=6pt, top=3pt, bottom=3pt, boxsep=0pt, before skip=0pt, after skip=0pt, fontupper=\small]\textcolor{xmarkred}{\ding{55}}~\textbf{Outcome.}\ #1\end{tcolorbox}}
\newenvironment{caselist}{\begin{list}{}{\setlength{\leftmargin}{3.4em}\setlength{\labelwidth}{3.0em}\setlength{\labelsep}{0.4em}\setlength{\itemsep}{1.5pt}\setlength{\parsep}{0pt}\setlength{\topsep}{2pt}\setlength{\partopsep}{0pt}}}{\end{list}}

This section describes how we annotate the failed attempts behind \cref{fig:belief}, defines its levels and behaviours, and gives a worked example of each.

\subsection{Annotation}
\label{sec:suppl:belief:annotation}

We derive the levels and behaviours below from a manual review of sampled reasoning traces and write them into the prompt in \cref{fig:suppl:belief_prompt}.
Claude Opus 5.5 then labels all 293 failed attempts of Kimi K3, Qwen3.8-Max, and Qwen3.8-27B at seed 36.
For each attempt, it receives the following inputs.
\begin{itemize}[leftmargin=1.5em, itemsep=1pt, topsep=3pt]
  \item \textbf{Reasoning}: the complete reasoning of the model and its visible replies, cut into numbered segments.
  \item \textbf{Actions}: every action that the model executes.
  \item \textbf{Root cause}: the cause of the failure under the taxonomy of \cref{sec:suppl:errors}, with its evidence.
  \item \textbf{Correct solution}: what is true, so that the judge knows what the model got wrong.
\end{itemize}
We cut the reasoning into numbered segments because it is long.
A failed attempt contains a median of 11.6k thinking tokens over six steps, and the longest one contains about 202k tokens.
Read as one block of text, such a log is hard to follow, and a judgement cannot be tied to a specific passage.
We therefore cut each step at paragraph, line, and sentence boundaries into segments of at most about 500 characters, which gives a median of 217 segments per attempt.
The numbers let the judge locate the passage in which the wrong belief first appears and cite the exact segments behind each judgement, which makes the labels more accurate and lets us check every quotation against the log.
The judge records the decisive wrong belief and the segment in which it first appears, the outcome of this belief, and whether each of the three behaviours occurs.
One author then checked the labels of 50 sampled attempts against the reasoning logs and agreed with 92\% of them.

\subsection{Levels of revising the wrong belief}
\label{sec:suppl:belief:levels}

For every failed attempt, we first fix the decisive wrong belief, that is, the wrong reading of the screen or the wrong assumption about the task that determines the failure.
The levels of \cref{fig:belief}a are nested, and an attempt counts toward a level when its reasoning reaches it.
\begin{itemize}[leftmargin=1.5em, itemsep=2pt, topsep=3pt]
  \item \textbf{Doubts the belief}: the reasoning states that the belief may be wrong, or the right alternative appears in it in any form.
  \item \textbf{Reasons out the right alternative}: the model states the correct answer or a largely complete solution as its own conclusion and supports it with an observation or a computation.
  A question, one untested option in a list, a correct candidate that the model eliminates on the basis of the wrong belief, and a finding that covers only a part of the solution do not count.
  \item \textbf{Gives up the wrong belief}: the model no longer holds the belief at the end of the attempt.
  \item \textbf{Corrects the belief}: the model ends with the correct view on the point that determines the failure.
\end{itemize}
Since the levels are nested, the share of attempts that stop at a level is the difference between this level and the next.
The outcome in the prompt is the highest level that an attempt reaches: \texttt{unshaken}, \texttt{doubted}, \texttt{reasoned\_out}, \texttt{replaced} (it gives up the belief for another wrong one), or \texttt{corrected}.
Cases~1--6 show attempts that stop at each level, from a belief never questioned to one corrected only after the attempt is decided.

\subsection{Behaviours that defend the wrong belief}
\label{sec:suppl:belief:behaviours}

The three behaviours of \cref{fig:belief}b are judged independently of the levels, and an attempt counts when the behaviour occurs at least once.
\begin{itemize}[leftmargin=1.5em, itemsep=2pt, topsep=3pt]
  \item \textbf{Fabricates an observation} (Case~7): after a wrong belief has formed, the model reports that it sees something on the screen that supports the belief, although the screenshot does not show it.
  The most frequent form is a second look that repeats the first wrong reading.
  \item \textbf{Decides by a hallucinated prior experience} (Case~8): the model asserts how puzzles of this kind are usually built, solved, or checked, although nothing on the screen states this, and uses the assertion to select, exclude, or confirm an answer.
  \item \textbf{Submits blindly as a probe} (Case~9): the model clicks Verify while unsure of its answer, to observe the response of the page or in the hope of a second chance.
\end{itemize}

\subsection{Reading the cases}
\label{sec:suppl:belief:cases}

Each case shows the task, the decisive wrong belief, what is in fact true, the model's screen beside the reference solution (in Case~6, beside the next screen), verbatim excerpts of the reasoning, and the outcome.
Coordinates in the excerpts are in the model's own frame, and rings or outlines on a screen mark what is selected on it.
The excerpts use three markings.
\begin{itemize}[leftmargin=1.5em, itemsep=1pt, topsep=3pt]
  \item \textcolor{casewrong}{Red} marks the wrong belief, evidence that is not on the screen, or an unsure submission.
  \item \textcolor{caseright}{Green} marks where the model states the right alternative.
  \item {\color{black!62}\itshape Grey notes} after $\hookrightarrow$ are ours.
\end{itemize}

\begin{casebox}{Case 1. Never questions the belief \textnormal{\textbar} Qwen3.8-27B}
\casetask{Tap every pumpkin in a 3D field scene that is mostly hidden behind another object; all the pumpkins are the same size.}
\begin{minipage}[t]{0.38\linewidth}\vspace{0pt}\centering\scriptsize
\begin{minipage}[t]{0.488\linewidth}\centering\includegraphics[width=\linewidth]{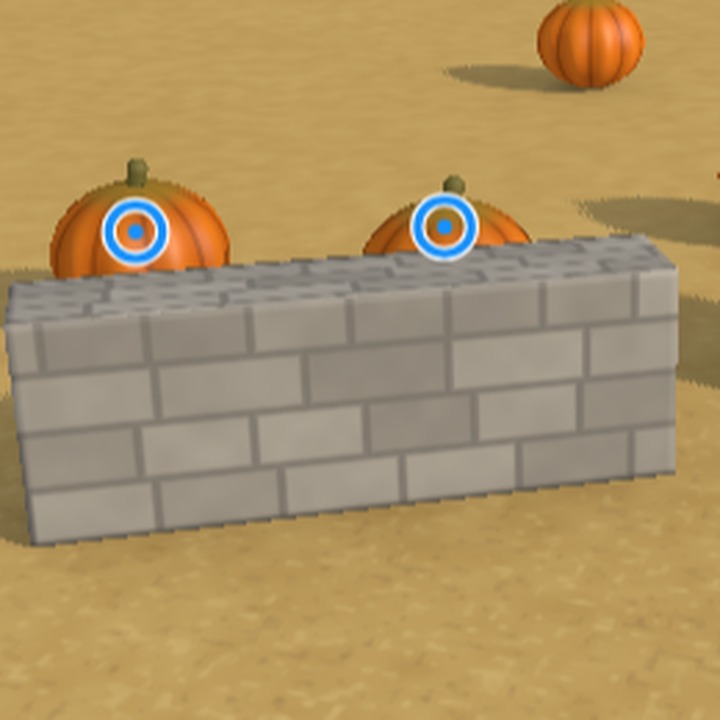}\\[1.5pt]Model's screen\end{minipage}\hfill
\begin{minipage}[t]{0.488\linewidth}\centering\includegraphics[width=\linewidth]{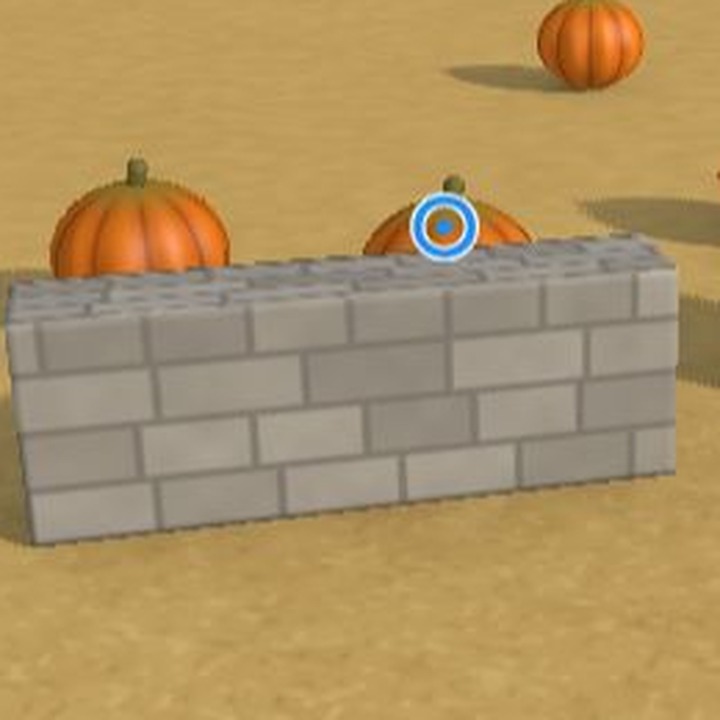}\\[1.5pt]Reference solution\end{minipage}
\end{minipage}\hfill
\begin{minipage}[t]{0.595\linewidth}\vspace{0pt}
\casebelief{The left pumpkin behind the wall is mostly hidden, because the wall covers the lower half of any pumpkin just behind it, and a hidden lower half counts as mostly hidden.}
\casetruth{The left wall pumpkin shows most of its body above the wall and is only about a third hidden; only the middle wall pumpkin and the pumpkin behind the stump are mostly hidden.}
\end{minipage}
\casereasoning
\begin{caselist}
\item[\casestep{1}] ``One on the left behind/near the brick wall (around 535, 298) - \textcolor{casewrong}{it's behind the brick wall, appears mostly hidden}''
\item[\casestep{2}] ``Actually, let me be careful. `Mostly hidden' means the bulk of the pumpkin is behind an object. The brick wall is tall enough to hide the bottom half. The pumpkins \textcolor{casewrong}{sitting just behind it have their lower half hidden = mostly hidden}.''\casenote{The self-check restates the rule instead of questioning it: half hidden is taken as mostly hidden.}
\item[\casestep{5}] ``The three I selected are \textcolor{casewrong}{the clearly hidden ones: two behind the brick wall} and one behind the stump. This seems correct. Let me verify.''
\end{caselist}
\caseoutcome{It submits the left wall pumpkin with the two mostly hidden ones, calling all three clearly hidden in its last check, so the selection is wrong.}
\end{casebox}

\begin{casebox}{Case 2. Only mentions the right alternative \textnormal{\textbar} Qwen3.8-Max}
\casetask{A photo of a living room is cut into nine tiles, some of them turned; tap tiles, a quarter turn per tap, until the whole photo is upright.}
\begin{minipage}[t]{0.38\linewidth}\vspace{0pt}\centering\scriptsize
\begin{minipage}[t]{0.488\linewidth}\centering\includegraphics[width=\linewidth]{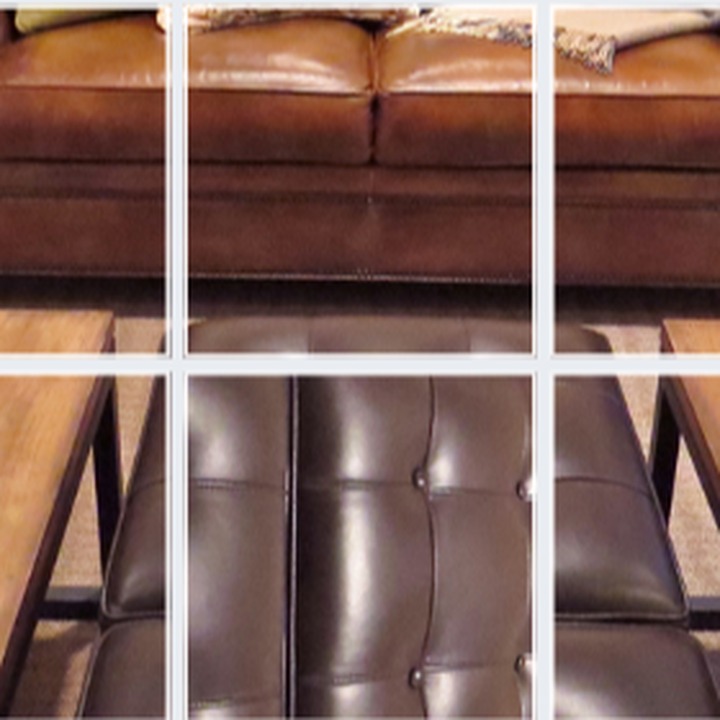}\\[1.5pt]Model's screen\end{minipage}\hfill
\begin{minipage}[t]{0.488\linewidth}\centering\includegraphics[width=\linewidth]{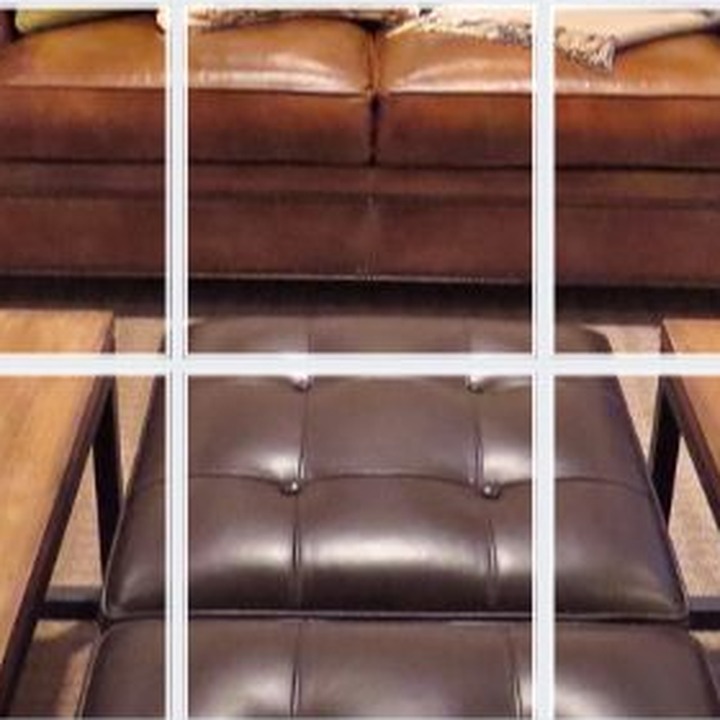}\\[1.5pt]Reference solution\end{minipage}
\end{minipage}\hfill
\begin{minipage}[t]{0.595\linewidth}\vspace{0pt}
\casebelief{The bottom row, including the bench tile, is already upright: although the bench seams run vertically, the model takes the orientation as ambiguous because such tufting is usually a grid.}
\casetruth{The bottom-middle tile, a tufted leather bench, is still turned a quarter turn: its seams break off at the tile borders. Upright, the long seams run horizontally and continue into the neighbouring tiles.}
\end{minipage}
\casereasoning
\begin{caselist}
\item[\casestep{1}] ``If the photo's foreground shows the bench from above, the bench's long axis is horizontal (parallel to sofa). \textcolor{caseright}{The tufted seams should run horizontally?}''
\item[\casestep{1}] ``Hmm, in the center tile the tufted lines run vertically (3 vertical columns of tufting). On a bench in front of the sofa, \textcolor{casewrong}{the tufting pattern is usually a grid; orientation ambiguous}.''
\item[\casestep{1}] ``Currently center tile shows vertical tufted columns (3 vertical bands). If bench's long axis is horizontal (parallel to sofa, spanning 3 tiles), then tufting grid: seams both directions; \textcolor{casewrong}{vertical seams visible within center tile are plausible upright}.''
\item[\casestep{3}] ``bottom row shows the ottoman/bench and floor. \textcolor{casewrong}{Everything appears upright and continuous.}''\casenote{In fact the bench seams in the middle tile do not line up with those in the tiles beside it.}
\end{caselist}
\caseoutcome{It turns the three top tiles upright, leaves the bench tile as it is and submits; eight of nine tiles are correct, so the attempt fails.}
\end{casebox}

\begin{casebox}{Case 3. Reasons out the right alternative, then rejects it \textnormal{\textbar} Kimi K3}
\casetask{Follow four arrow moves of a red crate atop stacked blocks, one square per arrow, falling straight down, and tap the block top where it lands.}
\begin{minipage}[t]{0.38\linewidth}\vspace{0pt}\centering\scriptsize
\begin{minipage}[t]{0.488\linewidth}\centering\includegraphics[width=\linewidth]{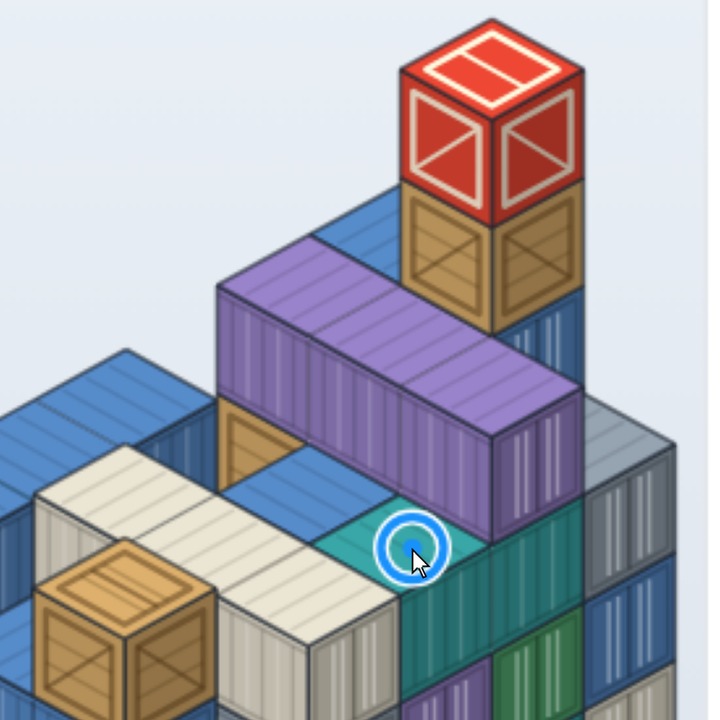}\\[1.5pt]Model's screen\end{minipage}\hfill
\begin{minipage}[t]{0.488\linewidth}\centering\includegraphics[width=\linewidth]{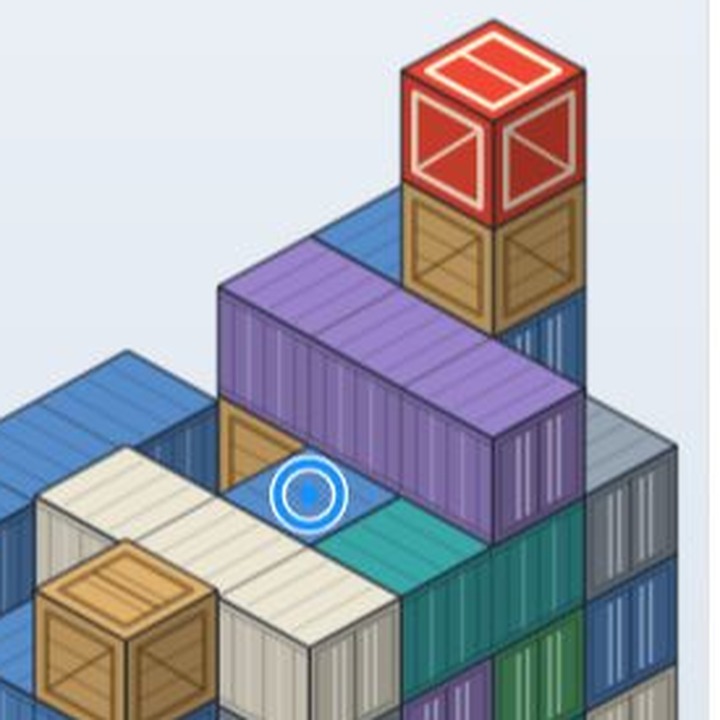}\\[1.5pt]Reference solution\end{minipage}
\end{minipage}\hfill
\begin{minipage}[t]{0.595\linewidth}\vspace{0pt}
\casebelief{The purple container is only two squares long, so the landing square is the teal top in front of its right end; it also places the blue top a level too low.}
\casetruth{The arrows add up to two squares down-left. The purple container is three squares long; the crate's third move ends on its middle square and the fourth drops it onto the blue top.}
\end{minipage}
\casereasoning
\begin{caselist}
\item[\casestep{15}] ``The purple block: \textcolor{casewrong}{it's clearly a 2x1 block (2 cells)}.''
\item[\casestep{20}] ``The spot 2$\swarrow$ from the red crate at the fallen height = \textcolor{casewrong}{the teal patch (in front of purple's right)}. I'm confident enough: TEAL.''
\item[\casestep{20}] ``\textcolor{caseright}{CONCLUSION FLIPS: target Q = C0+2b = the BLUE patch} (center $\approx$ 1331, 775), NOT teal!!!''\casenote{Q = C0+2b: the landing square, two squares down-left of the start. Its computed position falls on the blue top.}
\item[\casestep{20}] ``\textcolor{casewrong}{The blue patch = C0-a+2b = H-4} (from the brown-face argument). So Q $\ne$ blue (different levels!). Q (H-3) would be the TEAL (H-3)?? And teal = C0+2b \ding{51}\ding{51}\ding{51} \textcolor{casewrong}{TOPOLOGY RESTORED: teal = target!!}''\casenote{The blue top is in fact at H-3, one level below the purple, exactly where the crate lands.}
\end{caselist}
\caseoutcome{It taps the teal top and submits; the crate in fact lands on the blue top, one square up-left of the teal.}
\end{casebox}

\begin{casebox}{Case 4. Reasons out the right alternative, then rejects it \textnormal{\textbar} Qwen3.8-Max}
\casetask{In a 3D room with chairs, toy pickups and dogs on the floor, select the two objects that face the same direction in the room.}
\begin{minipage}[t]{0.38\linewidth}\vspace{0pt}\centering\scriptsize
\begin{minipage}[t]{0.488\linewidth}\centering\includegraphics[width=\linewidth]{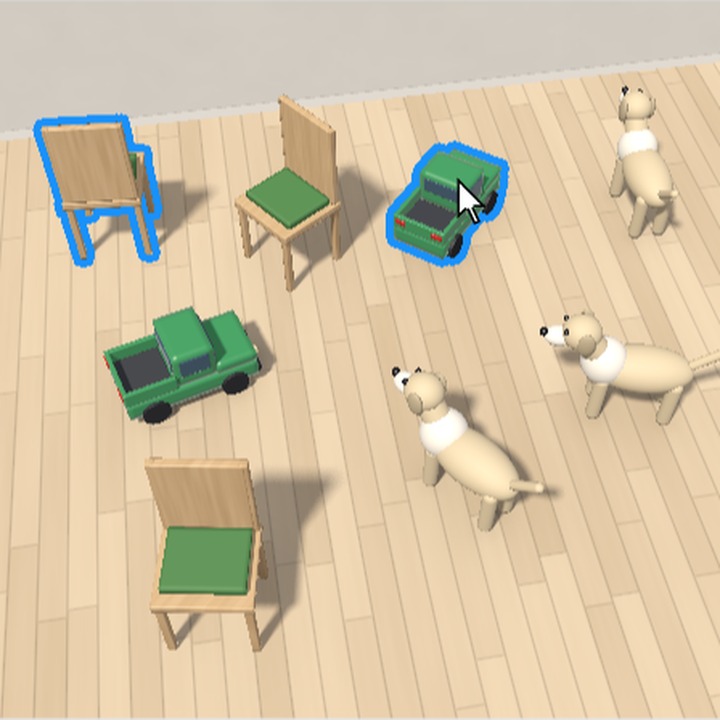}\\[1.5pt]Model's screen\end{minipage}\hfill
\begin{minipage}[t]{0.488\linewidth}\centering\includegraphics[width=\linewidth]{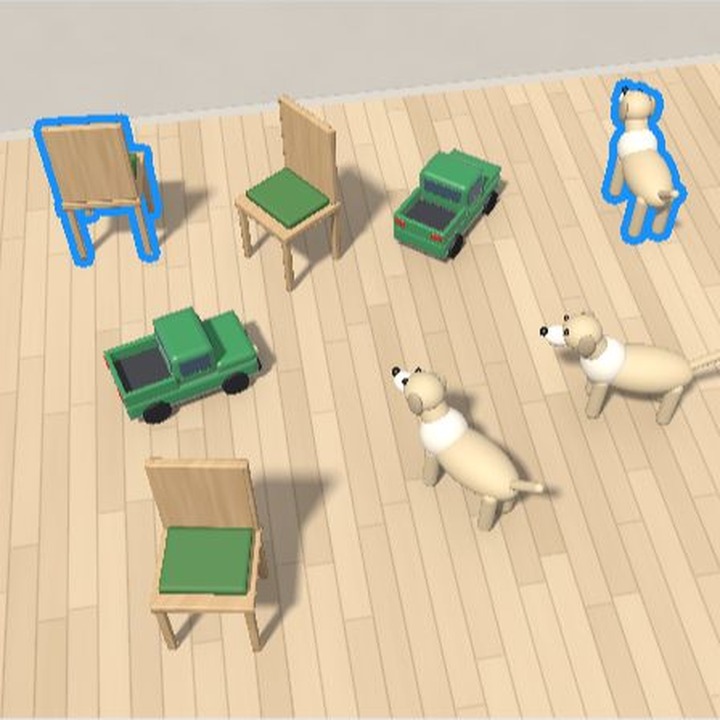}\\[1.5pt]Reference solution\end{minipage}
\end{minipage}\hfill
\begin{minipage}[t]{0.595\linewidth}\vspace{0pt}
\casebelief{Chair A is turned to face north-east, because its backrest looks narrower than the front chair's and a sliver of its seat shows on the right.}
\casetruth{Chair A (top left) and dog 1 (top right) both face the back wall, seen from behind; car 1, the pickup between them, points to the back right, about 36$^\circ$ away.}
\end{minipage}
\casereasoning
\begin{caselist}
\item[\casestep{1}] ``Hmm but \textcolor{caseright}{dog 1's head at top means facing north (toward wall)}.''
\item[\casestep{1}] ``If south, seat visible in front (below) of panel. \textcolor{caseright}{We don't see seat below panel; we see legs. So chair A faces north.}''
\item[\casestep{1}] ``\textcolor{caseright}{Thus answer: click chair A (\textasciitilde{}815, 375) and dog 1} (\textasciitilde{}1165, 370), then Verify.''
\item[\casestep{8}] ``\textcolor{casewrong}{Board width 52 vs 64: 80\% =\textgreater{} rotated =\textgreater{} NE.} Also the sliver of green on right supports seat offset to E =\textgreater{} NE (or E, but E board would be thin 35). \textcolor{casewrong}{So A = NE. Pair = (Chair A, Car 1).}''\casenote{Chair A is farther away than the front chair, so its backrest looks narrower; it still faces the wall.}
\end{caselist}
\caseoutcome{It selects chair A and car 1 and submits; the pair is wrong, because chair A's partner is dog 1, which it had found first.}
\end{casebox}

\begin{casebox}{Case 5. Replaces the belief with another wrong belief \textnormal{\textbar} Qwen3.8-27B}
\casetask{Stack cubes on a 3$\times$3 plan (bottom row marked FRONT) so that the back view and the right-side view match the two silhouettes shown.}
\begin{minipage}[t]{0.38\linewidth}\vspace{0pt}\centering\scriptsize
\begin{minipage}[t]{0.488\linewidth}\centering\includegraphics[width=\linewidth]{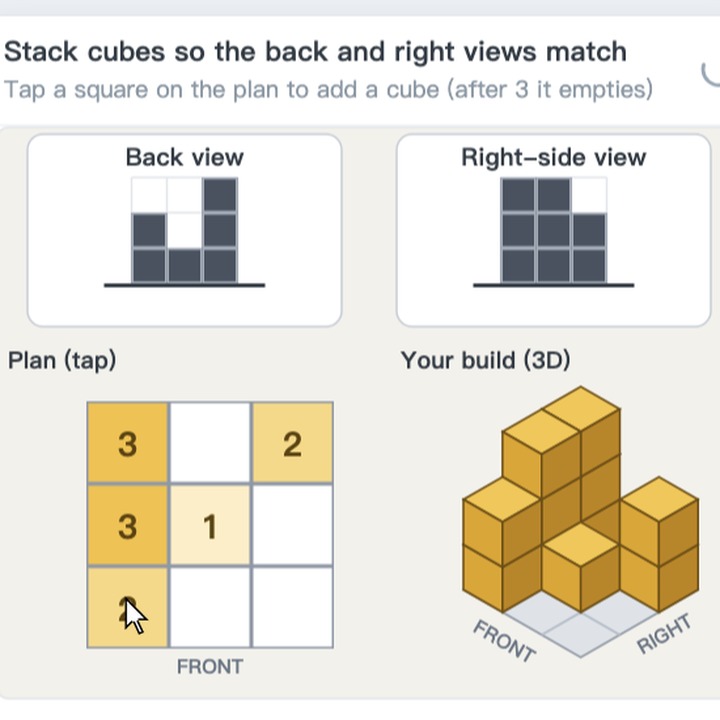}\\[1.5pt]Model's screen\end{minipage}\hfill
\begin{minipage}[t]{0.488\linewidth}\centering\includegraphics[width=\linewidth]{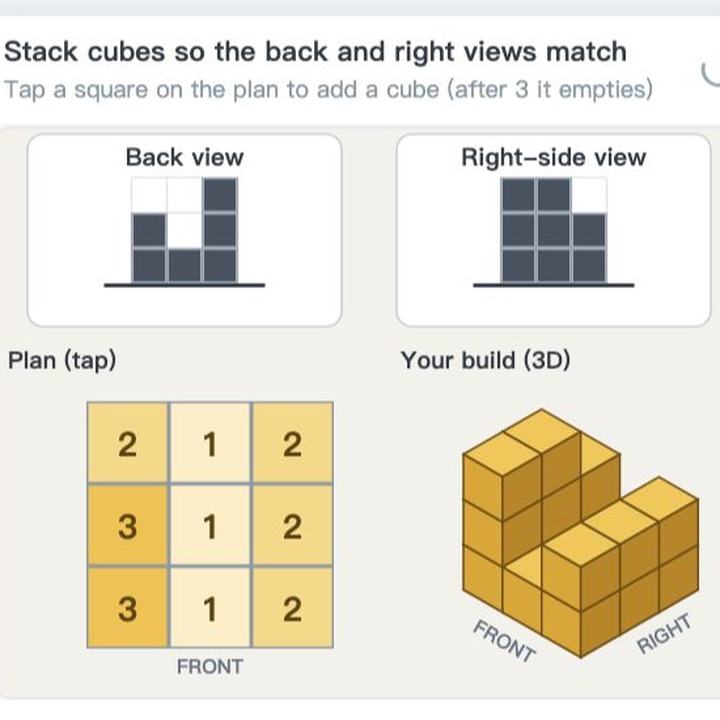}\\[1.5pt]Reference solution\end{minipage}
\end{minipage}\hfill
\begin{minipage}[t]{0.595\linewidth}\vspace{0pt}
\casebelief{First, the view's left column shows the back row. At the end, it shows the front row, but the plan's top row is now the front, so the build seems to fit.}
\casetruth{Seen from the right, the view's left column is the front (bottom) row and its right column the back (top) row, at most 2 high; the model's top row holds a stack of 3.}
\end{minipage}
\casereasoning
\begin{caselist}
\item[\casestep{2}] ``When looking from the right, you see the depth axis. \textcolor{casewrong}{The `left' in the right-side view corresponds to the back of the structure}, and `right' corresponds to the front.''
\item[\casestep{11}] ``If I'm facing the structure from the right side: - \textcolor{caseright}{The structure's FRONT is to my LEFT}''
\item[\casestep{11}] ``Let me recalculate: - \textcolor{casewrong}{Right view left (Row1 front) = max(3, 0, 2) = 3} - Right view middle (Row2) = max(3, 1, 0) = 3 - Right view right (Row3 back) = max(2, 0, 0) = 2''\casenote{Row 1 (3, empty, 2) is the plan's top row; FRONT is printed under the bottom row.}
\end{caselist}
\caseoutcome{It keeps the stack of 3 in the top (back) row and submits; the right-side view's right (back) column is 3, not 2.}
\end{casebox}

\begin{casebox}{Case 6. Corrects the belief, but too late \textnormal{\textbar} Kimi K3}
\casetask{A robot vacuum drives diagonally one tile per turn and bounces off walls; each turn, tap one toy into the basket or Wait; no toy may be swallowed.}
\begin{minipage}[t]{0.38\linewidth}\vspace{0pt}\centering\scriptsize
\begin{minipage}[t]{0.488\linewidth}\centering\includegraphics[width=\linewidth]{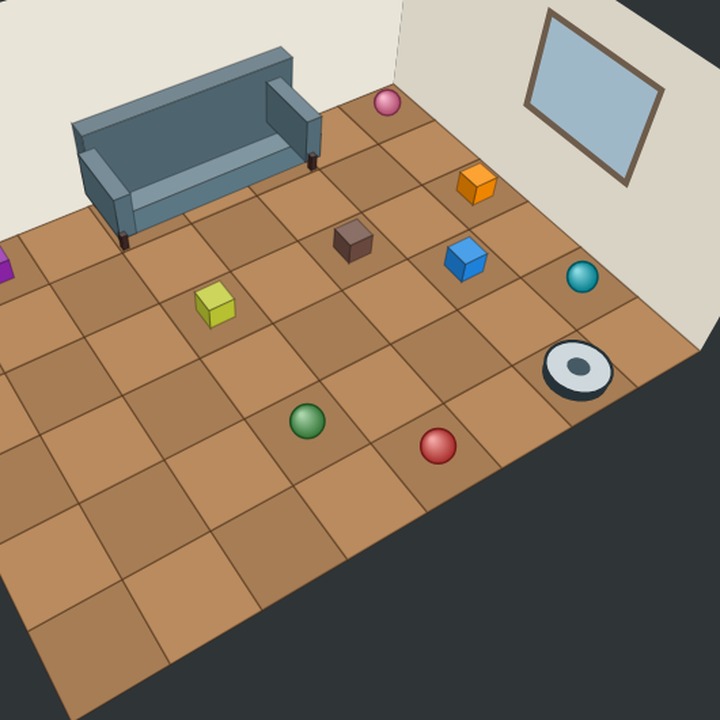}\\[1.5pt]Model's screen\end{minipage}\hfill
\begin{minipage}[t]{0.488\linewidth}\centering\includegraphics[width=\linewidth]{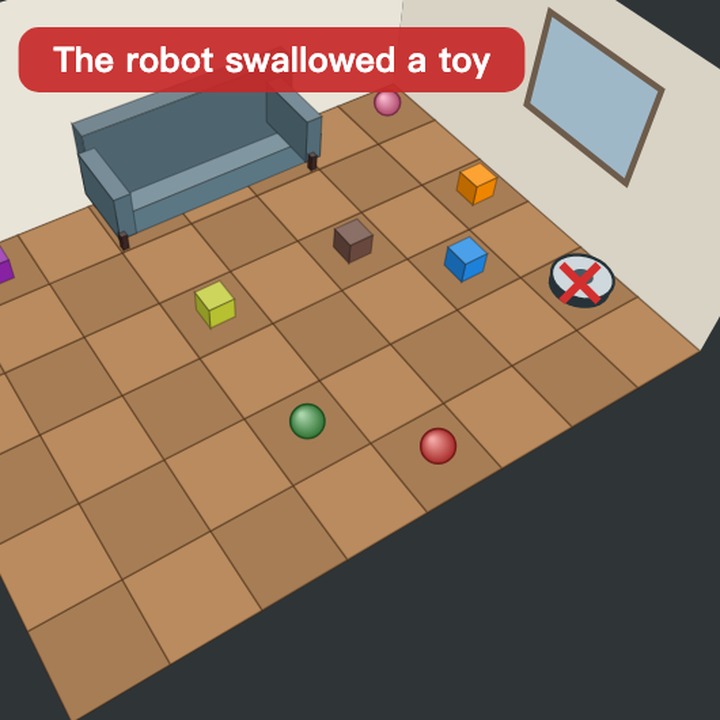}\\[1.5pt]Next screen, after Wait\end{minipage}
\end{minipage}\hfill
\begin{minipage}[t]{0.595\linewidth}\vspace{0pt}
\casebelief{The robot will make one more move to the right before it bounces, so the cyan ball, one tile above it, is safe this turn.}
\casetruth{The robot is already on the last tile before the floor's right edge, so its next move bounces it diagonally up onto the cyan ball's tile; the ball had to be basketed now.}
\end{minipage}
\casereasoning
\begin{caselist}
\item[\casestep{3}] ``They're at similar x but y differs by 68. That's about one `vertical' tile apart? The robot moves horizontally though, \textcolor{casewrong}{so it won't hit the cyan ball while moving right}.''
\item[\casestep{3}] ``\textcolor{casewrong}{It should bounce soon.} Let me observe the bounce behavior.''\casenote{It clicks Wait; the next screen shows the banner 'The robot swallowed a toy'.}
\item[\casestep{4}] ``This means my approach of just observing failed. \textcolor{caseright}{I needed to basket the cyan ball before the robot reached it.}''
\item[\casestep{4}] ``Actually maybe \textcolor{caseright}{the robot at turn 3 (1184,462) was already at the right wall}, and the `right' bounce reversed its horizontal grid component, changing direction from `right' to `up' (in screen).''
\end{caselist}
\caseoutcome{It clicks Wait; the robot bounces up onto the cyan ball and swallows it, which fails the puzzle before the model sees its mistake.}
\end{casebox}

\begin{casebox}{Case 7. Fabricates an observation to fit its belief \textnormal{\textbar} Qwen3.8-Max}
\casetask{In a flat maze of light paths between dark walls, click the one gem that the runner at the top can reach by following the open paths.}
\begin{minipage}[t]{0.38\linewidth}\vspace{0pt}\centering\scriptsize
\begin{minipage}[t]{0.488\linewidth}\centering\includegraphics[width=\linewidth]{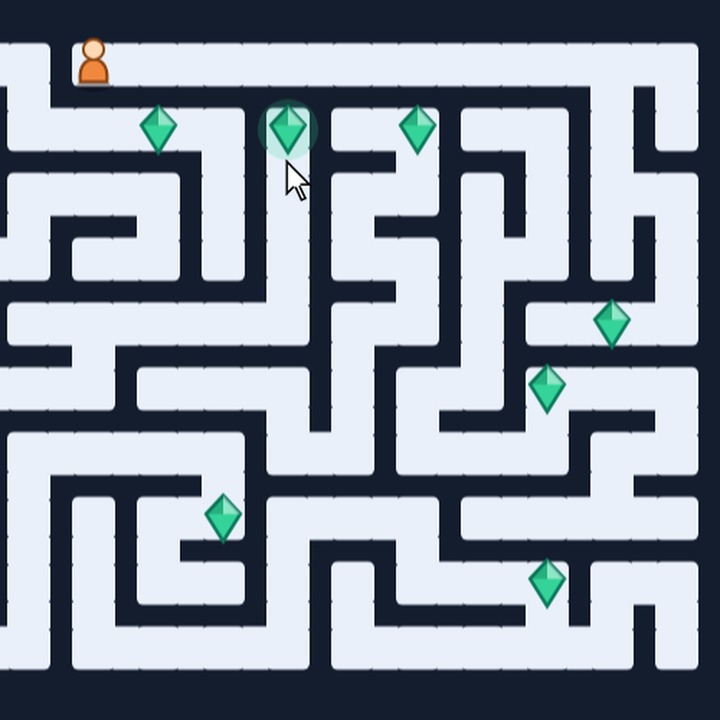}\\[1.5pt]Model's screen\end{minipage}\hfill
\begin{minipage}[t]{0.488\linewidth}\centering\includegraphics[width=\linewidth]{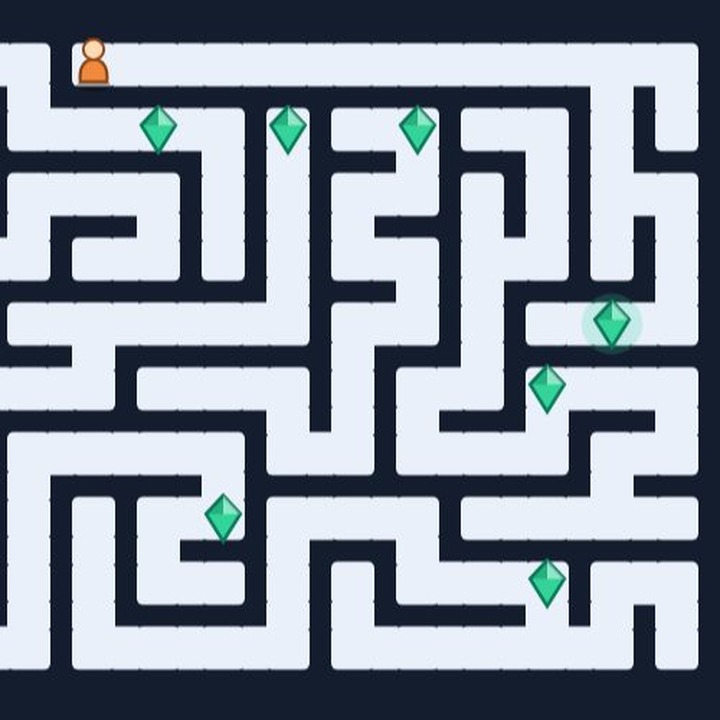}\\[1.5pt]Reference solution\end{minipage}
\end{minipage}\hfill
\begin{minipage}[t]{0.595\linewidth}\vspace{0pt}
\casebelief{The model believes that an open light passage runs straight down from the runner's corridor to the middle gem just below it, so that gem is reachable.}
\casetruth{An unbroken dark wall runs under the runner's top corridor; the corridor turns down only near the right edge, and that branch leads to the gem on the right.}
\end{minipage}
\casereasoning
\begin{caselist}
\item[\casestep{1}] ``I'm fairly confident because \textcolor{casewrong}{I see a continuous light vertical there crossing rows 1-2}.''\casenote{'There' is between the runner's corridor and the middle gem; the dark wall between them is unbroken.}
\item[\casestep{1}] ``And \textcolor{caseright}{the top-row gems' verticals are all blocked at row 1 (dark)}, making them isolated.''\casenote{Here it reads the walls correctly.}
\item[\casestep{1}] ``\textcolor{casewrong}{I see for G2 a clear vertical light from row 0 to row 3} (continuous).''\casenote{G2 is the middle gem; the model then commits to it.}
\item[\casestep{2}] ``It seems \textcolor{casewrong}{there's a continuous light from the top corridor to the gem}. Good.''\casenote{A second look, at a new screenshot, repeats the same reading.}
\end{caselist}
\caseoutcome{It selects the middle gem and clicks Verify; that gem is walled off from the runner, so the answer is wrong.}
\end{casebox}

\begin{casebox}{Case 8. Decides by hallucinated prior experience \textnormal{\textbar} Qwen3.8-Max}
\casetask{In a 3D scene of toys on a checkerboard floor, tap the three toys that stand directly behind the red cylinder, as seen from the camera.}
\begin{minipage}[t]{0.38\linewidth}\vspace{0pt}\centering\scriptsize
\begin{minipage}[t]{0.488\linewidth}\centering\includegraphics[width=\linewidth]{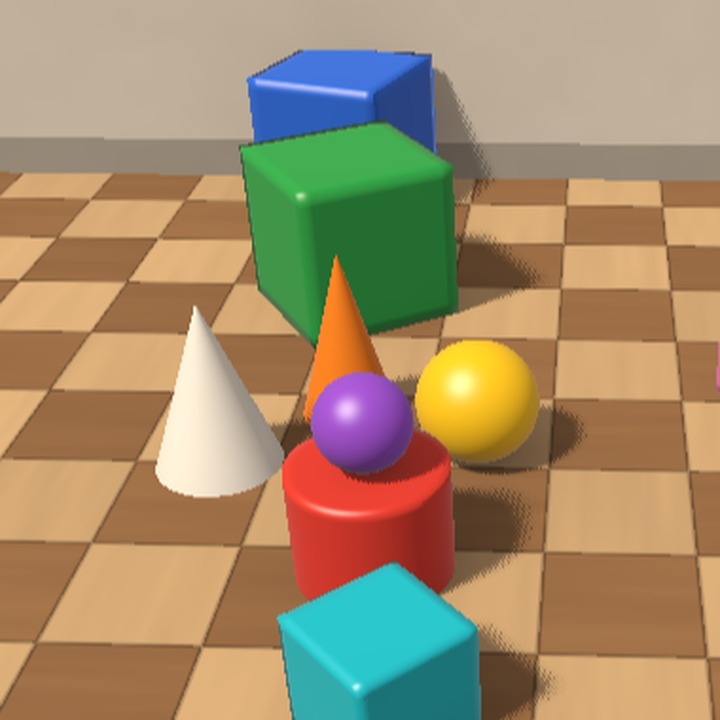}\\[1.5pt]Model's screen\end{minipage}\hfill
\begin{minipage}[t]{0.488\linewidth}\centering\includegraphics[width=\linewidth]{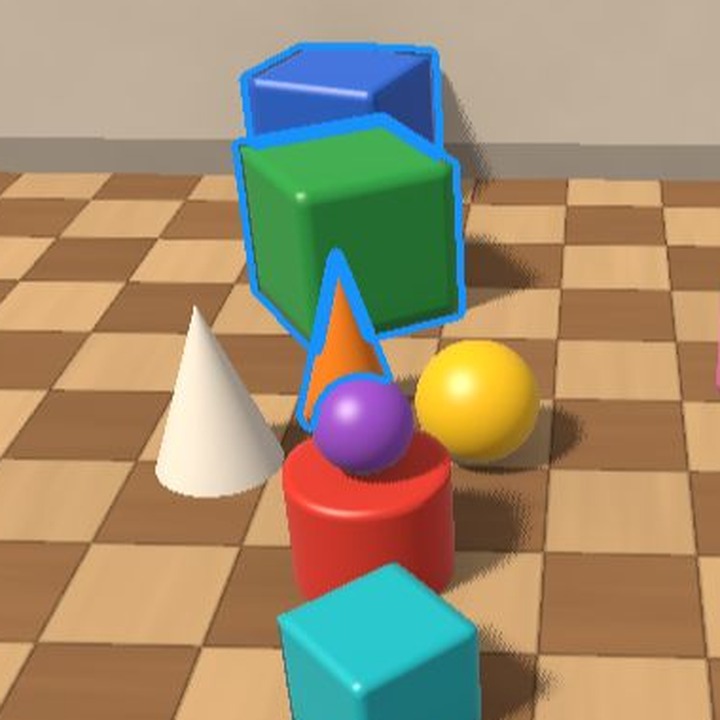}\\[1.5pt]Reference solution\end{minipage}
\end{minipage}\hfill
\begin{minipage}[t]{0.595\linewidth}\vspace{0pt}
\casebelief{The model takes the purple ball for a separate toy on the floor just behind the cylinder, because it assumes that toys are never stacked in such puzzles.}
\casetruth{The purple ball rests on top of the red cylinder, not behind it; the three toys behind the cylinder are the orange cone, the green cube and the blue cube.}
\end{minipage}
\casereasoning
\begin{caselist}
\item[\casestep{1}] ``Maybe \textcolor{caseright}{purple sphere is not behind but on top of cylinder}? If on top, then not `behind,' and \textcolor{caseright}{behind 3 = cone, green, blue}. Hmm! Look: purple sphere appears to rest on top of red cylinder?''
\item[\casestep{1}] ``\textcolor{casewrong}{Compare with typical generator: toys on floor only (no stacking)}. \textcolor{casewrong}{I've seen this CAPTCHA style}: `toybox' CAPTCHA with shapes on checkerboard; stacking not present. So purple on floor behind cylinder.''
\item[\casestep{1}] ``Unless purple sphere is on top of cylinder (not behind). \textcolor{caseright}{Then behind 3 = cone, green, blue. Yes! That resolves count!}''\casenote{Counting the ball, four toys line up behind the cylinder, one more than the three asked for.}
\item[\casestep{1}] ``\textcolor{casewrong}{I recall in these toy CAPTCHAs, stacking doesn't happen.} So purple on floor.''
\end{caselist}
\caseoutcome{It taps the purple ball, the orange cone and the green cube; the ball sits on the cylinder rather than behind it, so the answer is wrong.}
\end{casebox}

\begin{casebox}{Case 9. Submits blindly as a probe \textnormal{\textbar} Kimi K3}
\casetask{Panels show a square sheet folded twice, turned over and punched; tap every cell of the unfolded sheet that has a hole.}
\begin{minipage}[t]{0.38\linewidth}\vspace{0pt}\centering\scriptsize
\begin{minipage}[t]{0.488\linewidth}\centering\includegraphics[width=\linewidth]{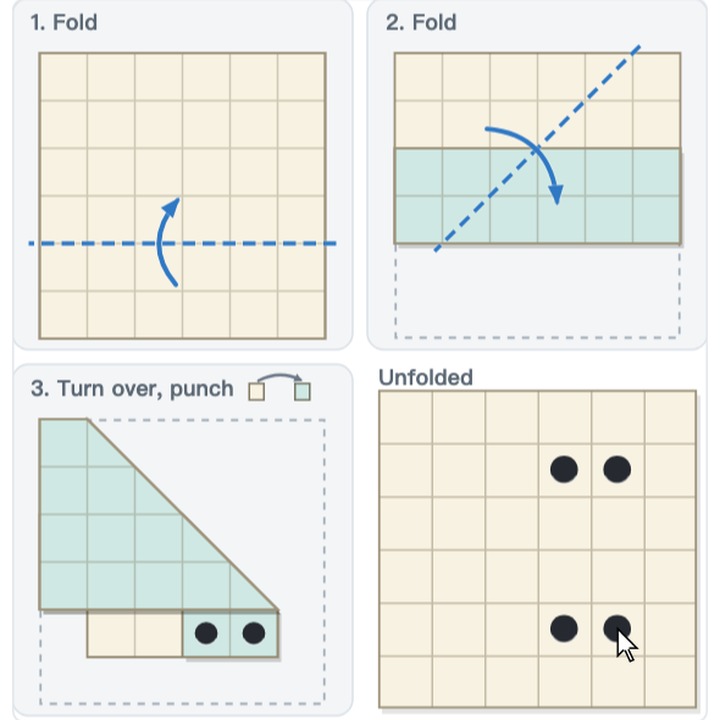}\\[1.5pt]Model's screen\end{minipage}\hfill
\begin{minipage}[t]{0.488\linewidth}\centering\includegraphics[width=\linewidth]{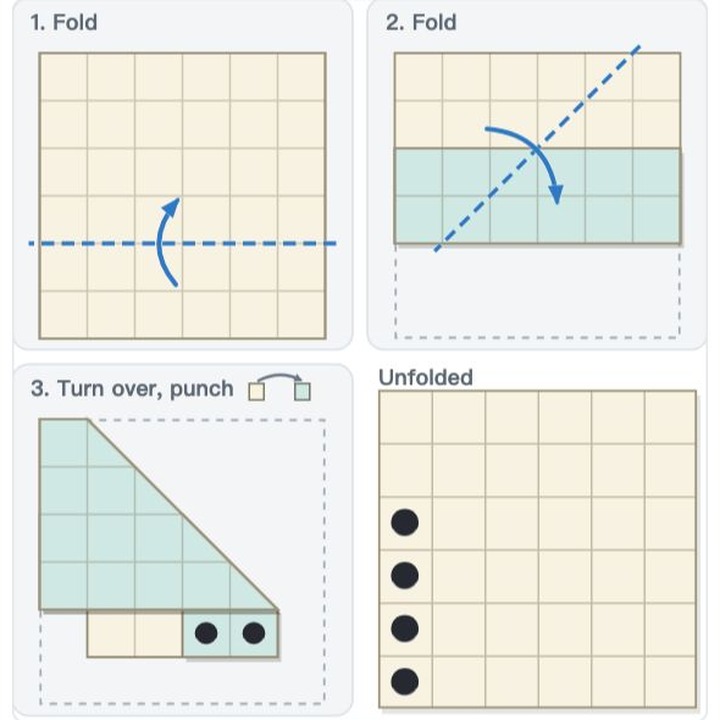}\\[1.5pt]Reference solution\end{minipage}
\end{minipage}\hfill
\begin{minipage}[t]{0.595\linewidth}\vspace{0pt}
\casebelief{The model believes the first fold runs along the middle of the sheet, between rows 3 and 4, so the bottom half folds onto the top half.}
\casetruth{The first fold line lies between rows 4 and 5, not at the middle, so only the bottom two rows fold up; the holes are in column 1, rows 3 to 6.}
\end{minipage}
\casereasoning
\begin{caselist}
\item[\casestep{1}] ``Panel 1 (Fold 1): Full 6x6 square. \textcolor{casewrong}{Horizontal dashed line at middle (between row 3 and row 4)}.''\casenote{Panel 1 has four rows above the dashed line and two below it.}
\item[\casestep{1}] ``So fold1 line is at 2/3, folding the bottom 1/3 up? Arrow up: \textcolor{caseright}{bottom third (rows 5--6) folds up onto rows 3--4}.''
\item[\casestep{1}] ``Actually, I realize \textcolor{casewrong}{I cannot resolve fine geometry from text reasoning reliably}. The expected behavior: click the correct cells then Verify. \textcolor{casewrong}{If wrong, it likely resets and I can retry} (there's a refresh icon).''
\item[\casestep{1}] ``Actually I already have a clear screenshot. \textcolor{casewrong}{Let me just commit to a best-effort and if it fails use refresh}.''
\end{caselist}
\caseoutcome{Unsure, it marks columns 4 and 5 in rows 2 and 5 and clicks Verify; none of these is a hole, and the attempt is checked only once.}
\end{casebox}

\clearpage
\section{Error analysis}
\label{sec:suppl:errors}

This section details the failure analysis in \cref{sec:analysis:failures}.

\subsection{Labelling procedure}
\label{sec:suppl:errors:procedure}

We analyze all 380 failed attempts of the three strongest models, \mbox{GPT-6} Astra, Claude Opus 5.5, and Gemini 4 Argon, and assign each of them one root cause.
From a manual review of sampled trajectories, we derive ten causes in three groups: perception, reasoning, and interaction.

The root cause is the earliest error that decides the outcome, so that the attempt would most likely have passed without it.
To keep the causes disjoint, the label follows a fixed order of checks.
If the first decisive error is a wrong reading of a screenshot that the model had already received, the cause is in perception; otherwise, if the model read the scene correctly but drew the wrong conclusion, it is in reasoning; otherwise it is in interaction.
The rules at the end of the prompt settle the boundary cases, such as a model that accepts a visible mismatch in its final state (perception) against one that never obtained the screenshot showing the mismatch (missed observation).
Claude Opus 5.5 serves as an agent-as-a-judge: with the prompt in \cref{fig:suppl:prompt}, it reads the model's reasoning when the API returns it, every action, the screenshots, the checker result, the final screenshot of the oracle, and the task program, which it can run offline to replay the attempt, and assigns exactly one cause.
One author then independently labels all 380 failures again, and the two sets of labels match on 90.3\% of the failures (Cohen's $\kappa=0.88$), which confirms the judge's reliability.
\Cref{tab:errors} reports the human labels as final and omits grounding error, the tenth cause, since it never occurs.

\begin{suppblock}
  \captionsetup{type=table} %
  \footnotesize
  \setlength{\tabcolsep}{2.5pt}
  \captionof{table}{Failure taxonomy of the 380 failed attempts of the three strongest models. Group rows give the number and share of failures, and the Astra, Opus, and Argon columns count the failures of each model.}
  \label{tab:errors}
  \resizebox{\linewidth}{!}{%
  \begin{tabular}{@{}l>{\raggedright\arraybackslash}p{4.6cm}rccc>{\raggedright\arraybackslash}p{6.2cm}@{}}
    \toprule
    Error cause & What goes wrong & \#Fail. & Astra & Opus & Argon & Representative example \\
    \midrule
    \rowcolor{keybg}
    \multicolumn{7}{c}{\emph{Perception} (217, 57.1\%)} \\
    Depth misjudgment     & Misjudges 3D size, distance, occlusion, or pose in the flat image & 152 & 40 & 60 & 52 & Asked to rank glasses by water held, puts the nearest first because it merely looks largest \\
    Object misrecognition & Overlooks, miscounts, or confuses visible objects within one view & 36 & 2 & 13 & 21 & Asked to restore a picture by swapping tiles, checks it but leaves two similar tiles misplaced \\
    Measurement error     & Misjudges a visible angle, length, or offset, with no depth involved & 29 & 7 & 7 & 15 & Asked to straighten a crooked photo, checks it, judges it level, and submits it still 3$^\circ$ off \\
    \rowcolor{keybg}
    \multicolumn{7}{c}{\emph{Reasoning} (135, 35.5\%)} \\
    Rotation error        & Imagines a rotation, fold, cut, or mirror image wrongly & 33 & 7 & 5 & 21 & Asked which cubes a lettered net folds into, mentally folds a letter the wrong way \\
    Viewpoint confusion   & Keeps the camera's view, or fails to match objects across cameras & 17 & 4 & 4 & 9 & Asked how many people three cameras show, counts a person seen by two cameras twice \\
    Motion misprediction  & Sees every key frame but mispredicts how things will move & 34 & 10 & 11 & 13 & Asked to cross turning planks, watches them, then steps onto one that turns away mid-step \\
    Wrong criterion       & Reads the 3D layout correctly but tests a simpler property than asked & 39 & 6 & 19 & 14 & Asked what a rabbit can hide behind, picks large objects, never checking the space behind \\
    Planning error        & Orders its own moves into a dead end or wastes its limited moves & 12 & 6 & 4 & 2 & Asked to push two crates onto marks, first pushes one so it blocks the other's only path \\
    \rowcolor{keybg}
    \multicolumn{7}{c}{\emph{Interaction} (28, 7.4\%)} \\
    Missed observation    & Skips a frame holding a key cue, by batching actions or acting too early & 28 & 15 & 7 & 6 & Asked which pair meets eyes, sends a batch of nine frame steps and skips the moment they do \\
    \bottomrule
  \end{tabular}}
\end{suppblock}

\subsection{Findings}
\label{sec:suppl:errors:findings}

\noindent\textbf{Frontier MLLMs act on the GUI accurately.}
Interaction is the smallest group, with only 7.4\% of the failures.
It contains no grounding error, so every click and drag lands where the model meant it to, and all 28 of its failures are missed observations, in which the model skips a key frame by batching actions or acting too early.
Acting on the screen is therefore not what limits frontier MLLMs on our tasks.

\medskip\noindent\textbf{Most failures come from misreading the screenshot.}
Perception is the largest group, with 57.1\% of failures, and depth misjudgment is the leading cause of every model, at 34--46\% of its failures.
The remaining 35.5\% spread over five reasoning causes, the largest of which, wrong criterion, decides 39 failures, against 152 for depth misjudgment.
Reading the 3D scene behind a flat screenshot is thus the main bottleneck of frontier MLLMs.

\medskip\noindent\textbf{Misreading also decides failures on tasks built to test reasoning.}
Mental transformation tasks require rotating, folding, or cutting an object in the mind, yet perception errors cause 35 of their 63 failures.
A model that misreads the scene has nothing correct to transform, so these attempts fail before any transformation begins.

\medskip\noindent\textbf{Visual feedback does not correct misreading.}
Closed-loop manipulation and planning tasks show a new screenshot after every action, so a model can check the result of each move.
Yet perception errors still cause 85 of their 129 failures, including 48 of the 63 on closed-loop manipulation.
For example, a model straightens a crooked photo, checks the result, judges it level, and submits it still 3$^\circ$ off.
Frontier MLLMs therefore misread even the result of their own actions.

\medskip\noindent\textbf{A stronger model mainly misreads less, whereas predicting motion is equally hard for all three.}
Gemini 4 Argon fails 153 of its 375 attempts and \mbox{GPT-6} Astra only 97, and perception errors account for 39 of these 56 additional failures.
Motion misprediction, in contrast, hardly differs, with 10, 11, and 13 failures for \mbox{GPT-6} Astra, Claude Opus 5.5, and Gemini 4 Argon.
The strongest model thus leads mainly by reading the screenshot better, not by predicting motion better.

\clearpage
\section{Comparison with related benchmarks}
\label{sec:suppl:compare}

\Cref{tab:captcha_compare} compares \bench{} with the CAPTCHA benchmarks and the interactive spatial benchmarks discussed in \cref{sec:related} along two groups of factors.

\subsection{Factors}
\label{sec:suppl:compare:factors}

The first group consists of four properties that describe how a model interacts with the tasks.
\begin{itemize}[leftmargin=1.5em, itemsep=2pt, topsep=3pt]
  \item \textbf{GUI actions}: the model acts only through pixel-level mouse and keyboard actions on a screenshot, as a \mbox{computer-use agent does}.
  \item \textbf{Observable}: every task type can be solved from static screenshots, without timing, motion, or trajectory checks.
  \item \textbf{Multi-turn}: at least one task type takes several rounds of acting, observing the result, and adjusting the next action.
  \item \textbf{Scalable}: every task type generates new instances on demand.
\end{itemize}
The second group consists of the five spatial categories of \bench{} (\cref{fig:teaser}, top).
A benchmark is marked for a category when at least one of its task types tests it.

\begin{suppblock}
  \captionsetup{type=table} %
  \small
  \setlength{\tabcolsep}{2pt}
  \captionof{table}{Comparison with CAPTCHA benchmarks and interactive spatial benchmarks. The factors are defined in \cref{sec:suppl:compare:factors}, and \cmark{} marks a factor that a benchmark satisfies or a spatial category that one of its task types tests.}
  \label{tab:captcha_compare}
  \newcommand{\hd}[1]{\shortstack{#1}}
  \newcommand{\vn}[1]{{\footnotesize #1}}
  \resizebox{\linewidth}{!}{%
  \begin{tabular}{@{}llcccccccccc@{}}
    \toprule
    & & & & & & \multicolumn{5}{c}{Spatial Intelligence} & \\
    \cmidrule(lr){7-11}
    Benchmark & Venue & \hd{GUI\\actions} & \hd{Obser-\\vable} & \hd{Multi-\\turn} & \hd{Scal-\\able} & \hd{Spatial\\relation} & \hd{Closed-loop\\manipulation} & \hd{Planning \&\\simulation} & \hd{Mental\\transform.} & \hd{Spatio-\\temporal} & \hd{\#Task\\types} \\
    \midrule
    \multicolumn{12}{@{}l}{\emph{CAPTCHA benchmarks}} \\
    MCA-Bench~\citep{wu2026mcabench} & \vn{AAAI'26} & \xmark & \xmark & \xmark & \xmark & \xmark & \xmark & \xmark & \xmark & \xmark & 20 \\
    Open CaptchaWorld~\citep{luo2025opencaptchaworld} & \vn{NeurIPS'25} & \xmark & \xmark & \cmark & \xmark & \xmark & \cmark & \xmark & \xmark & \xmark & 19 \\
    Next-Gen CAPTCHAs~\citep{liu2026nextgencaptchas} & \vn{ICML'26} & \xmark & \xmark & \cmark & \xmark & \xmark & \xmark & \cmark & \cmark & \cmark & 27 \\
    Halligan~\citep{teoh2025halligan} & \vn{USENIX'25} & \xmark & \cmark & \cmark & \xmark & \xmark & \xmark & \xmark & \xmark & \xmark & 26 \\
    Spatial CAPTCHA~\citep{kharlamova2026spatialcaptcha} & \vn{ICLR'26} & \xmark & \cmark & \xmark & \cmark & \cmark & \xmark & \xmark & \cmark & \xmark & 7 \\
    ReCAP~\citep{chen2026recap} & \vn{ICML'26} & \cmark & \cmark & \cmark & \cmark & \xmark & \xmark & \xmark & \xmark & \xmark & 7 \\
    CaptchaMind~\citep{wang2026captchamind} & \vn{EMNLP'26} & \xmark & \cmark & \cmark & \cmark & \xmark & \xmark & \xmark & \xmark & \xmark & 8 \\
    \midrule
    \multicolumn{12}{@{}l}{\emph{Interactive spatial benchmarks}} \\
    iVISPAR~\citep{mayer2025ivispar} & \vn{EMNLP'25} & \xmark & \cmark & \cmark & \cmark & \xmark & \cmark & \cmark & \xmark & \xmark & 1 \\
    VisGym~\citep{wang2026visgym} & \vn{NeurIPS'26} & \xmark & \cmark & \cmark & \cmark & \xmark & \cmark & \cmark & \xmark & \xmark & 17 \\
    ESI-Bench~\citep{hong2026esibenchembodiedspatialintelligence} & \vn{Preprint} & \xmark & \cmark & \cmark & \xmark & \cmark & \xmark & \cmark & \xmark & \cmark & 29 \\
    \midrule
    \bench{} (ours) & \vn{--} & \cmark & \cmark & \cmark & \cmark & \cmark & \cmark & \cmark & \cmark & \cmark & 125 \\
    \bottomrule
  \end{tabular}}
\end{suppblock}

\subsection{Differences from \bench{}}
\label{sec:suppl:compare:differences}

\noindent\textbf{CAPTCHA benchmarks.}
MCA-Bench, Halligan, ReCAP, and CaptchaMind consist of conventional web CAPTCHAs, such as distorted text, image grids, sliders, and icon matching, which require no spatial understanding.
The other three benchmarks cover some of the five spatial categories, but the model interacts with them differently from \bench{}:
\begin{itemize}[leftmargin=1.5em, itemsep=2pt, topsep=3pt]
  \item \textbf{Open CaptchaWorld} and \textbf{Next-Gen CAPTCHAs} contain task types that rely on timing, motion, or trajectory checks, and they do not generate new instances for every task type.
  \item \textbf{Spatial CAPTCHA} asks multiple-choice questions about static images, so the model gives a single answer and never acts on the scene.
\end{itemize}
Among the seven benchmarks, only ReCAP lets the model act through GUI actions as in \bench{}.

\medskip\noindent\textbf{Interactive spatial benchmarks.}
iVISPAR, VisGym, and ESI-Bench test spatial abilities over several turns, in a sliding puzzle, in 17 visual environments, and in simulated embodied scenes, respectively.
In all three, the model acts through text commands that are specific to each task rather than through GUI actions.
None of them tests mental transformation: the rotation tasks of VisGym display the result of every rotation, so they test closed-loop manipulation instead.

\medskip\noindent
In contrast, \bench{} poses every task as an interactive CAPTCHA on a web page, where the model acts only through GUI actions and observes each result in the next screenshot.
It has all four interaction properties and covers all five spatial categories with 125 task types, more than four times as many as any other benchmark in the table.

\clearpage
\section{Evaluation prompts}
\label{sec:suppl:eval_prompts}

This section gives the evaluation prompts of \cref{sec:exp:protocol}.
Each model runs in its official agent, which puts the same instruction, ``Solve the CAPTCHA puzzle shown on the screen. Follow the instruction written on the puzzle, then click Verify to submit your answer.'', into its own user-message template with the screenshot.

The system prompt is also each agent's official one.
We remove only the sentences that do not apply to a puzzle page: those about the OSWorld virtual machine, its password, home directory, and date, about files, applications, and dialogs, about asking the user, and about infeasible tasks.
UI-Venus-2-9B also loses its rule against a final submit click, which would forbid pressing Verify.
We then append one sentence that authorizes the model to solve the puzzles, shown in \textcolor{keyblue}{blue} below.
The Claude and Gemini models have no official system prompt, so this sentence is their whole system prompt.
ReCAP-32B-Instruct uses its authors' prompts.

\Cref{fig:suppl:eval_gpt,fig:suppl:eval_claude,fig:suppl:eval_gemini,fig:suppl:eval_kimi,fig:suppl:eval_qwen,fig:suppl:eval_evocua,fig:suppl:eval_opencua,fig:suppl:eval_venus,fig:suppl:eval_uitars,fig:suppl:eval_recap} show each system prompt with the first user message, in which {\color{black!50}\fontfamily{lmtt}\selectfont\itshape [screenshot]} marks the screenshot.
Tools defined through the API, such as the computer-use tools of the GPT, Claude, and Gemini models, are not shown.
\modelname receives the messages of its training samples in \cref{fig:suppl:train_system,fig:suppl:train_sample}.
When a model stops to ask whether it may solve the puzzle, the attempt continues with the reply in \cref{fig:suppl:eval_claude}.

\par\begingroup\captionsetup{type=figure}%
\begin{promptbox}{Evaluation prompts of GPT-6 Astra and GPT-6 Sol}
  \promptpart{System prompt}{eval_gpt_system}
  \promptpart{First user message}{eval_gpt_user}
\end{promptbox}\nopagebreak\vspace{9pt}\nopagebreak
\captionof{figure}{System prompt and first user message of the GPT-6 models.}
\label{fig:suppl:eval_gpt}
\par\endgroup\medskip

\par\begingroup\captionsetup{type=figure}%
\begin{promptbox}{Evaluation prompts of Claude Opus 5.5, Claude Fable 5.1, and Claude Sonnet 5.5}
  \promptpart{System prompt}{eval_auth_system}
  \promptpart{First user message}{eval_claude_user}
  \promptpart{Reply when a model stops to ask, sent with the current screenshot}{eval_claude_continue}
\end{promptbox}\nopagebreak\vspace{9pt}\nopagebreak
\captionof{figure}{System prompt, first user message, and reply to a model that stops to ask, for the Claude models.}
\label{fig:suppl:eval_claude}
\par\endgroup\medskip

\par\begingroup\captionsetup{type=figure}%
\begin{promptbox}{Evaluation prompts of Gemini 4 Argon and Gemini 3.8 Flash}
  \promptpart{System prompt}{eval_auth_system}
  \promptpart{First user message}{eval_gemini_user}
\end{promptbox}\nopagebreak\vspace{9pt}\nopagebreak
\captionof{figure}{System prompt and first user message of the Gemini models.}
\label{fig:suppl:eval_gemini}
\par\endgroup\medskip

\par\begingroup\captionsetup{type=figure}%
\begin{promptbox}{Evaluation prompts of Kimi K3}
  \promptpart{System prompt}{eval_kimi_system}
  \promptpart{First user message}{eval_kimi_user}
  \promptpart{Assistant turn written by the agent}{eval_kimi_assistant}
  \promptpart{Tool result}{eval_kimi_tool}
\end{promptbox}\nopagebreak\vspace{9pt}\nopagebreak
\captionof{figure}{System prompt and first turns of Kimi K3, including the screenshot turn that its agent writes.}
\label{fig:suppl:eval_kimi}
\par\endgroup\medskip

\par\begingroup\captionsetup{type=figure}%
\begin{promptbox}[breakable]{Evaluation prompts of Qwen3.5-9B, Qwen3.8-27B, and Qwen3.8-Max}
  \promptpart{System prompt}{eval_qwen_system}
  \promptpart{First user message}{eval_qwen_user}
\end{promptbox}\nopagebreak\vspace{9pt}\nopagebreak
\captionof{figure}{System prompt and first user message of the Qwen models.}
\label{fig:suppl:eval_qwen}
\par\endgroup\medskip

\par\begingroup\captionsetup{type=figure}%
\begin{promptbox}[breakable]{Evaluation prompts of EvoCUA-32B}
  \promptpart{System prompt}{eval_evocua_system}
  \promptpart{First user message}{eval_evocua_user}
\end{promptbox}\nopagebreak\vspace{9pt}\nopagebreak
\captionof{figure}{System prompt and first user message of EvoCUA-32B.}
\label{fig:suppl:eval_evocua}
\par\endgroup\medskip

\par\begingroup\captionsetup{type=figure}%
\begin{promptbox}{Evaluation prompts of OpenCUA-32B}
  \promptpart{System prompt}{eval_opencua_system}
  \promptpart{First user message}{eval_opencua_user}
\end{promptbox}\nopagebreak\vspace{9pt}\nopagebreak
\captionof{figure}{System prompt and first user message of OpenCUA-32B.}
\label{fig:suppl:eval_opencua}
\par\endgroup\medskip

\par\begingroup\captionsetup{type=figure}%
\begin{promptbox}[breakable]{Evaluation prompts of UI-Venus-2-9B}
  \promptpart{System prompt}{eval_venus_system}
  \promptpart{First user message}{eval_venus_user}
\end{promptbox}\nopagebreak\vspace{9pt}\nopagebreak
\captionof{figure}{System prompt and first user message of UI-Venus-2-9B.}
\label{fig:suppl:eval_venus}
\par\endgroup\medskip

\par\begingroup\captionsetup{type=figure}%
\enlargethispage{2\baselineskip}
\begin{promptbox}{Evaluation prompts of UI-TARS-1.5-7B}
  \promptpart{System prompt}{eval_uitars_system}
  \promptpart{First user message}{eval_uitars_user}
  \promptpart{Second user message}{eval_uitars_user2}
\end{promptbox}\nopagebreak\vspace{9pt}\nopagebreak
\captionof{figure}{System prompt and first two user messages of UI-TARS-1.5-7B.}
\label{fig:suppl:eval_uitars}
\par\endgroup\medskip

\par\begingroup\captionsetup{type=figure}%
\begin{promptbox}[breakable]{Evaluation prompt of ReCAP-32B-Instruct}
  \promptpart{First user message (no system prompt)}{eval_recap_user}
\end{promptbox}\nopagebreak\vspace{9pt}\nopagebreak
\captionof{figure}{First user message of ReCAP-32B-Instruct, which receives no system prompt.}
\label{fig:suppl:eval_recap}
\par\endgroup\medskip

\clearpage
\section{Annotation prompts}
\label{sec:suppl:prompt}

\Cref{fig:suppl:belief_prompt} gives the prompt for the wrong-belief analysis of \cref{sec:analysis:belief} (\cref{sec:suppl:belief}), and \cref{fig:suppl:prompt} the prompt with which an agent assigns the root cause of a failed attempt for \cref{sec:analysis:failures} (\cref{sec:suppl:errors}).

\par\begingroup\captionsetup{type=figure}%
\begin{promptbox}[breakable]{Wrong-belief annotation prompt of Claude Opus 5.5}
  \fontsize{7.0}{8.35}\selectfont
  You are analyzing a failed attempt of a multimodal model on an interactive spatial puzzle. The model sees the puzzle only through screenshots, acts with mouse and keyboard actions, and finally clicks Verify to submit. A submission is checked once, and the model receives no feedback afterwards.

  \smallskip
  You receive a reading file with two parts. The \emph{header} gives the root cause of the failure, the evidence for it, and the correct solution; treat it as given, since it establishes what the model got wrong and what is true. The \emph{trajectory} gives, for every step, the model's reasoning and its visible reply, cut into numbered segments (\texttt{[s3\_12]} is reasoning segment 12 of step 3, and \texttt{[r3\_1]} the first segment of the reply of step 3), followed by the executed action. The text of a step was written while the model viewed the screenshot named for that step. A line marked \texttt{TOOL ERROR} was not shown to the model. The model may reason in its own coordinate frame, which is not an error in itself. Very long logs are abridged, and every omission is marked. Base every judgement on what the model wrote, and record the following.

  \smallskip
  \textbf{1. Decisive wrong belief.} The single wrong belief that determined the failure: a wrong reading of the screen, a wrong rule, or a wrong assumption about the task or the page. If several beliefs contributed, take the one that the root cause identifies. Record the segment in which it first appears, and one sentence on what the model believes and what is actually true.

  \smallskip
  \textbf{2. Outcome of the belief.} Assign the first of the following outcomes that applies.
  \begin{itemize}[leftmargin=1.2em,itemsep=0pt,topsep=1pt]
    \item \texttt{corrected}: by the end of the log, the model has abandoned the belief and holds the correct view on the point that determined the failure, by stating it or by acting on it.
    \item \texttt{replaced}: the model abandons the belief but ends with a different wrong reading or rule on the same point.
    \item \texttt{reasoned\_out}: the model keeps the belief to the end, but at some point it reasons out the correct alternative. A passage must state the correct answer or a largely complete solution \emph{specifically} (the object, value, direction, order, or move that is correct, not a method or an unspecific possibility), \emph{ground} it in an observation, a measurement, a trace, a computation, or an argument, or adopt it as the current answer or act on it, and \emph{assert} it as the model's own result or view, even briefly. A passage written before the belief has settled counts.
    \item \texttt{doubted}: the model keeps the belief and never reasons out the correct alternative, but it questions the belief: it states that it may have misread, miscounted, or assumed wrongly on this point, or the correct alternative appears in a weaker form, such as a question, a conditional, one untested option in a list, a candidate that the model eliminates on the basis of the wrong belief, or a correct finding on only a part of the point.
    \item \texttt{unshaken}: none of the above applies. Doubt about anything other than the decisive belief, a general principle that the model never applies to the case, and a shift between variants of the same error do not count.
  \end{itemize}
  \textbf{3. Behaviours.} For each behaviour, cite one segment that shows it, or record \texttt{null} if it does not occur. Judge the behaviours independently of one another and of the outcome.
  \begin{itemize}[leftmargin=1.2em,itemsep=0pt,topsep=1pt]
    \item \texttt{fabricated\_observation}: after a wrong belief or expectation has appeared, the model reports as seen (``I can see'', ``the image shows'', ``in the zoomed view'', or a description of a new screenshot) something that supports it but that the screen does not show. A second look that repeats the wrong reading counts. The passage in which the belief first appears, an inference, a restatement that does not claim to see anything, and a stated uncertainty do not. Examine every later step before you record \texttt{null}.
    \item \texttt{hallucinated\_rule}: the model asserts how puzzles of this kind are usually built, solved, or checked, such as the usual number or form of the answers, typical layouts, orientations, or distractors, or the criterion of the checker, and uses this assertion to select, exclude, or confirm an answer, although nothing on the screen states it. An assumption about retries, a guess without an asserted rule, correct general knowledge, and a remark that is not used for the answer do not count.
    \item \texttt{blind\_submit}: the model clicks Verify while it states that it is unsure of its answer, in order to observe the response of the page, or counting on a second chance, that is, it states anywhere in the log that a wrong submission can be retried, resets the puzzle, or returns feedback, or what it will do if the submission fails. An attempt without Verify, exploratory actions other than Verify, and a submission made with stated confidence do not count.
  \end{itemize}
  Answer with one JSON object and nothing else:\\
  \texttt{\{"belief": \{"segment", "statement"\},}\\
  \texttt{\ "outcome": "corrected" | "replaced" | "reasoned\_out" | "doubted" | "unshaken",}\\
  \texttt{\ "evidence":} one or two segments that decide the outcome, each with a verbatim quotation of at most 160 characters,\\
  \texttt{\ "behaviours": \{"fabricated\_observation", "hallucinated\_rule", "blind\_submit"\}:} each \texttt{null} or a segment with a verbatim quotation,\\
  \texttt{\ "rationale":} one sentence that justifies the outcome\texttt{\}}
\end{promptbox}\nopagebreak\vspace{9pt}\nopagebreak
\captionof{figure}{Prompt with which Claude Opus 5.5 annotates the decisive wrong belief, its outcome, and the three behaviours of a failed attempt (\cref{sec:suppl:belief}).}
\label{fig:suppl:belief_prompt}
\par\endgroup\medskip

\clearpage
\par\begingroup\captionsetup{type=figure}%
\begin{promptbox}[breakable]{Root-cause annotation prompt of Claude Opus 5.5}
  \fontsize{7.0}{8.35}\selectfont
  You are analyzing why a multimodal model failed a task in TuringSpace. Each task is an interactive spatial CAPTCHA on a web page. The model plays it as a computer-use agent: it receives a screenshot, issues mouse or keyboard actions, receives the next screenshot, and finally presses Verify to submit. Some tasks show a clip frame by frame (a Next button advances one frame) or advance one frame per move. A checker decides whether the final state is correct.

  \smallskip
  You receive one failed attempt: the instruction, every model call with the model's visible reasoning (full text, a summary, or a short intent per action, depending on the model), every action with the screenshot it produced, the checker's result, the final screenshot of a correct solution, and the program of the task, which you may run to regenerate the instance and replay the actions.

  \smallskip
  Find the root cause of the failure: the single error that decided the outcome, i.e.\ the error without which the attempt would most likely have passed. If there are several, take the earliest one. Assign exactly one of the following labels.

  \smallskip
  \textbf{Perception}: the model misreads what the screenshots show.
  \begin{itemize}[leftmargin=1.2em,itemsep=0pt,topsep=1pt]
    \item \emph{Depth misjudgment.} It reads distance, height, or occlusion off the 2D screen: it compares on-screen sizes, distances, or positions without correcting for depth or height, treats objects that overlap or touch on the screen as touching in 3D, misjudges which object is in front, or misses a tilt toward or away from the camera. Examples: picks the cube nearest the camera as the largest, although it only looks the largest; steps between two blocks that only touch on the screen.
    \item \emph{Object misrecognition.} Within a single view, it misses, miscounts, or confuses a visible object, part, marking, or color, including one that is still wrong in its own final state. Examples: takes two people in near-identical shirts for one; declares a picture puzzle solved while two look-alike tiles are still swapped.
    \item \emph{Measurement error.} It reads the layout correctly and measures the right quantity with the right method but misestimates its value for a reason other than perspective, such as an angle, a length, or an offset, and accepts the result after looking at it. Example: straightens a crooked photo but submits it still slightly crooked.
  \end{itemize}
  \textbf{Reasoning}: the model reads the scene correctly but draws the wrong conclusion.
  \begin{itemize}[leftmargin=1.2em,itemsep=0pt,topsep=1pt]
    \item \emph{Rotation error.} It imagines the wrong result of rotating, folding, mirroring, or cutting an object. Examples: accepts the mirror image of the target shape; folds a net with one letter turned the wrong way.
    \item \emph{Viewpoint confusion.} It keeps the camera's view where the task asks for another view, such as a person's own front, left, and right, a view from the back or the side, or the ground rather than a moving vehicle, or it fails to match the same object across two views or cameras. Examples: takes the camera's left for the left of a person who faces the camera; counts a person seen by two cameras twice.
    \item \emph{Motion misprediction.} It has seen every frame available before acting but predicts wrongly how things move or change over time, such as where a ball rolls, where a beam goes after a mirror, when a swinging load comes to rest once no wait is left, or where a moving object will be on the next frame. Example: steps onto a turning plank that is aligned now but has turned away when the step lands.
    \item \emph{Wrong criterion.} It checks an easier property than the condition that the task asks for, such as size or shape instead of whether an object fits, passes, opens, or hides. Examples: picks every large solid object as a hiding place without checking whether there is room behind it; picks only the openings with exactly the shape of a tile, not every opening the tile can drop through.
    \item \emph{Planning error.} It plans its own moves badly: in an order that leads to a dead end, or by wasting moves when the number of moves is limited. Example: pushes the first crate where it blocks the second.
  \end{itemize}
  \textbf{Interaction}: the reasoning is right, but the model does not observe the screen it needs or does not execute what it intended.
  \begin{itemize}[leftmargin=1.2em,itemsep=0pt,topsep=1pt]
    \item \emph{Missed observation.} It commits although waiting or stepping one frame at a time would still have shown the screen that decides the answer: it advances several frames in one call, makes an irreversible move before the scene shows what it needs to know, or submits without looking at the result of its last action. Example: sends a batch of nine frame steps and skips the frames in which two people make eye contact.
    \item \emph{Grounding error.} The model names the right target, but its action misses it: the click lands just beside the element that the model named, or a drag over- or undershoots the value that the model stated.
  \end{itemize}
  \textbf{Others}: none of the labels above fits, or the attempt ended for a reason outside the model, such as an API call that never returned or a provider that refused the request. Describe the cause in one sentence.

  \smallskip
  Rules for close cases:
  \begin{itemize}[leftmargin=1.2em,itemsep=0pt,topsep=1pt]
    \item If the model looked at a screenshot that shows the problem and still judged it wrong, the cause is in Perception or Reasoning. If it never saw that screenshot because of how it acted, the cause is Missed observation.
    \item Batching actions or skipping frames is not by itself a missed observation. If the screenshots that the model did see already show the cue that decides the answer, the cause is in Perception or Reasoning.
    \item A click that lands where the model meant to click is not a grounding error, even if that place is wrong. Label the error that made the model choose that place.
    \item If the model miscounts or misidentifies objects because it misreads which object is in front, which is higher, or which touches which, use Depth misjudgment. An angle that is wrong because of foreshortening is also Depth misjudgment; any other wrong angle is Measurement error.
    \item If the model's own description of the 3D layout is wrong and this error leads to its wrong action, use Depth misjudgment even if it then also applies a simple rule. A wrong remark that does not change what the model does is not the root cause. If the layout is right and the condition of the task is replaced by an easier property, or a condition that the task states is never checked, use Wrong criterion.
    \item If the model picks a place or a direction from where things lie on the screen, such as the spot directly below a person or a shape that only looks right on the screen, instead of from their 3D positions, use Depth misjudgment. Use Viewpoint confusion when the model uses the camera's left, right, front, or back for someone or something that faces another way, or fails to match the same object across two views.
    \item If a rotation happens on the screen and the model misjudges what it can see, use a Perception label: Object misrecognition for a tile or piece left in the wrong orientation, Measurement error for an angle that is slightly off. Use Rotation error only for a rotation, fold, or mirror image that the model has to imagine.
    \item If the model turns or tilts the board itself, use Rotation error when it mispredicts how the board ends up, and Motion misprediction when the turn is right but it mispredicts how things then move.
    \item If the error is in how objects in the scene move by themselves, use Motion misprediction. If it is in the order of the model's own moves, use Planning error.
    \item If the model gives no reasoning, choose the label that best explains its actions and its final answer. A wrong answer that is exactly what a flat reading of the screenshot predicts is Depth misjudgment.
  \end{itemize}
  Answer with one JSON object and nothing else:\\
  \texttt{\{"group": "Perception" | "Reasoning" | "Interaction" | "Others",}\\
  \texttt{\ "root\_cause":} the name of one cause above, such as \texttt{"Depth misjudgment"}, or \texttt{"Others"},\\
  \texttt{\ "evidence":} one sentence of concrete evidence: what the model did and what was correct\texttt{\}}
\end{promptbox}\nopagebreak\vspace{9pt}\nopagebreak
\captionof{figure}{Prompt with which an agent assigns the root cause of a failed attempt in \cref{tab:errors}.}
\label{fig:suppl:prompt}
\par\endgroup\medskip

\clearpage
\section{Ethics statement}
\label{sec:suppl:ethics}

\bench{} studies the spatial understanding of MLLMs through interactive CAPTCHAs.
The goal of this work is to measure and improve this understanding, not to bypass the CAPTCHAs that protect real websites.
As with other work on CAPTCHA solving, this work is dual-use.
Agents trained on \gymname{} may also solve deployed CAPTCHAs more often.
Our analysis identifies the spatial abilities that current agents lack.
These findings can guide the design of CAPTCHAs that stay easy for humans and hard for agents.
We request that \benchname{} not be used to train models.

\end{document}